\documentclass{article}

\usepackage[preprint]{neurips_2026}

\usepackage[utf8]{inputenc} 
\usepackage[T1]{fontenc}
\usepackage{url} 
\usepackage{booktabs} 
\usepackage{amsfonts} 
\usepackage{natbib}
\usepackage{colortbl}
\usepackage{arydshln}
\usepackage{booktabs}
\usepackage{nicefrac}  
\usepackage{microtype}  
\usepackage{tcolorbox}
\usepackage{tabularx}
\usepackage{xcolor}
\usepackage{amsmath}
\usepackage{graphicx}
\usepackage{subfigure}
\usepackage{subcaption}
\usepackage{wrapfig}
\usepackage{fontawesome5}
\usepackage{amssymb}
\usepackage{array}
\usepackage{caption}
\usepackage{pifont}
\usepackage{multirow}
\usepackage{color}

\usepackage[table,dvipsnames,svgnames]{xcolor}
\usepackage[table]{xcolor}
\usepackage{listings} 
\lstdefinestyle{mystyle}{
  language=Python,
  basicstyle=\ttfamily\footnotesize,
  backgroundcolor=\color{gray!10},
  frame=single,
  breaklines=true,
  showstringspaces=false
}
\usepackage[dvipsnames]{xcolor}
\usepackage{makecell}
\definecolor{mydarkred}{rgb}{0.6,0,0}
\definecolor{mydarkgreen}{rgb}{0,0.6,0}
\PassOptionsToPackage{options}{natbib}
\setcitestyle{authoryear,round,citesep={;},aysep={,},yysep={;}}
\definecolor{mydarkblue}{RGB}{70,130,180}
\usepackage[colorlinks,
linkcolor=mydarkred,
citecolor=mydarkblue,
urlcolor=mydarkblue]{hyperref}

\usepackage{algorithm}
\usepackage[noend]{algorithmic}

\definecolor{done}{RGB}{0, 150, 0}  
\definecolor{inprog}{RGB}{255, 140, 0} 
\definecolor{todo}{RGB}{200, 0, 0}   
\definecolor{note}{RGB}{100, 100, 200}  
\setcitestyle{numbers,square}
\definecolor{sapphireblue}{HTML}{0F52BA}
\definecolor{deepsapphire}{HTML}{082567}
\definecolor{sapphirebg}{HTML}{F2F6FF}

\title{AgentBrew: Offline Tool-Use Agent Learning from Raw Real-World Trajectories}

\author{%
  Zhiyi Lyu\textsuperscript{1} \And
  Yewen Li\textsuperscript{2} \And
  Longtao Zheng\textsuperscript{1} \And
  Shengtian Yang\textsuperscript{3} \And
  Lang Feng\textsuperscript{1} \And
  Lei Feng\textsuperscript{3} \And
  Peng Jiang\textsuperscript{2} \And
  Kun Gai\textsuperscript{2} \And
  Qingpeng Cai\textsuperscript{2} \And
  Bo An\textsuperscript{1} \\
  \AND
  \textnormal{\textsuperscript{1} Nanyang Technological University \quad
  \textsuperscript{2} Kuaishou Technology \quad
  \textsuperscript{3} Southeast University}
}

\begin{document}
\renewcommand{\algorithmicrequire}{\textbf{Input:}}
\renewcommand{\algorithmicensure}{\textbf{Output:}}

\maketitle

\begin{abstract}
LLM-based agents are increasingly deployed in real-world applications through tool-use APIs, yet training them for specific environments remains fundamentally difficult: real-world applications provide no pre-defined tasks or verifiers, no faithful simulators, and limited budget for large-scale environment interaction.
In this paper, we propose \textbf{AgentBrew}, an offline training framework that learns effective tool-use policies from a single batch of raw interaction trajectories, without task verifiers or iterative on-policy rollouts. The agent first explores the target environment to collect a raw trajectory corpus without quality filtering. To extract training signal from this noisy corpus, \emph{retrospective task inference} reconstructs an aligned instruction for each trajectory based on its actual outcome, and \emph{PMI-Based credit assignment} decomposes the trajectory's total information about the inferred instruction into additive per-action credits via pointwise mutual information (PMI). These credits weight the policy training objective, amplifying informative actions while suppressing ineffective ones.
On three real-world MCP applications (GitHub, Notion, PostgreSQL), AgentBrew improves Qwen3-32B by +8.7 Acc / +9.7 Score on average, surpassing Qwen3-235B (+2.3 / +4.4) and outperforming rejection sampling (+5.9 / +10.3). These results demonstrate that fine-grained offline learning can recover useful supervision from raw trajectories that filtering-based approaches would discard. The code is available at \url{https://github.com/alphatogo/AgentBrew}
\end{abstract}

\section{Introduction}
\label{sec: Introduction}
Tool-use agents powered by Large Language Models (LLMs) are rapidly moving from research prototypes to production deployments \citep{react, reloc, toolformer, mcp}. Standardized protocols like the Model Context Protocol (MCP) \citep{mcp} now connect LLM agents to real-world applications such as Notion \citep{notion_ai} and GitHub \citep{github_mcp_server}, enabling tool-use across complex, stateful environments. However, each application presents a distinct tool landscape: Notion exposes dozens of APIs spanning database operations, page management, and content editing, each requiring precise parameter composition and awareness of the application's live state. General-purpose LLMs, even at frontier scale, struggle with such environment-specific complexity, exhibiting low task completion rates on real-world tool-use benchmarks \citep{mcp-universe, mcpmark}, indicating that reliable deployment demands targeted, environment-specific training.

Existing approaches to training tool-use agents for specific applications follow two main routes. The first operates on real-world data \citep{toucan,xlam,agentdpo}: practitioners design a suite of tasks with ground-truth verifiers for the target application, deploy an agent to interact with the live environment and collect trajectories, then use the verifiers to filter for successful completions and train via supervised fine-tuning on the retained high-quality subset \citep{toolllm, xlam, apigen-mt}. The second pursues on-policy optimization \citep{grpo, gigpo, hehierarchy}: because repeated rollouts against live APIs incur monetary costs, rate limits, and irreversible side effects on production data, these methods instead construct simulated environments that replicate the application's state and tool behavior, enabling large-scale iterative training before deploying the resulting policy back to the real application \citep{agentworld, awm, scaleenv,androidworld}. Despite their different trade-offs, both routes rest on shared prerequisites: pre-defined task suites, reliable reward signals, and either sufficient real-world interaction budget or a faithful simulator of the target environment.

 \begin{wrapfigure}{r}{0.45\textwidth}
  \centering
  \includegraphics[width=0.45\textwidth]{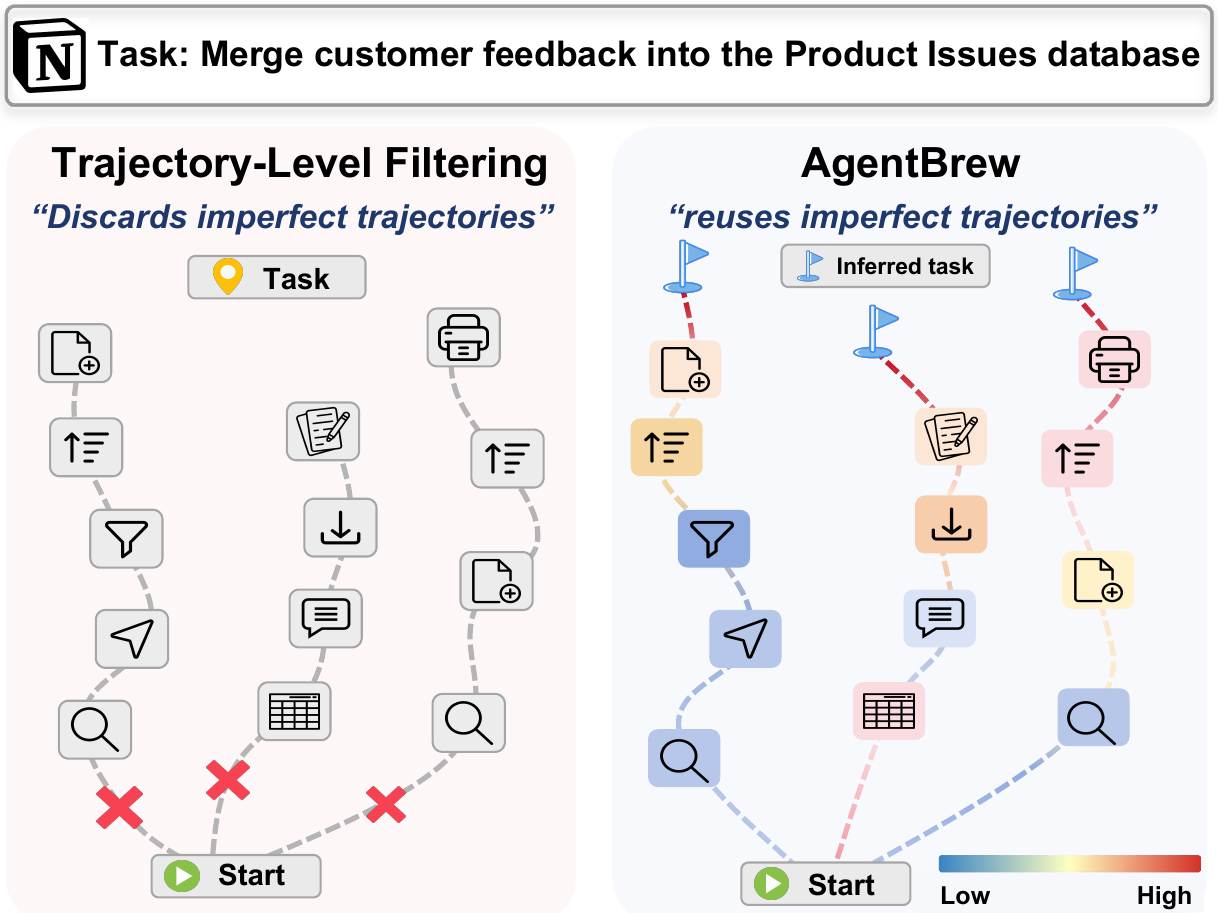}
  \caption{Left: trajectory-level filtering discards entire trajectories that fail the original task, losing potentially useful actions. Right: AgentBrew infers a better-aligned task for each trajectory and assigns per-action credits under the inferred task, enabling the entire corpus to contribute to training.}
  \label{fig:wrap}
  \vspace{-3mm}
\end{wrapfigure}

While these methods have achieved strong results on curated benchmarks, applying them to real-world applications exposes fundamental difficulties at three levels.
At the \textbf{task level}, real-world applications provide neither pre-defined task nor ground-truth verifiers. Tasks must be constructed from scratch with validity dependent on the environment's live state, and reliable verification remains an open problem \citep{stabletoolbench}: neither manual annotation \citep{workarena, androidworld} nor automated judging scales to complex, stateful API interactions (see \S\ref{sec:main_results}).
At the \textbf{trajectory level}, the offline route faces a severe data efficiency problem. Real-world tool-use tasks involve long, complex API compositions with low completion rates, so standard trajectory-level filtering \citep{toucan, xlam} discards the majority of an already limited corpus. Worse, even a 'failed' trajectory may contain actions that are useful training signal for a different but related task, yet trajectory-level filtering discards them entirely (Figure~\ref{fig:wrap}, left).
At the \textbf{training level}, the online route introduces a significant sim-to-real gap: simulators cannot faithfully reproduce the complex, interconnected state of real-world applications. Recent work confirms this empirically: simulator-trained agents achieve strong results on synthetic benchmarks yet degrade when evaluated on real-world platforms \citep{agentworld}.
Together, these challenges rule out standard training pipelines and point to a setting that, despite being the norm for industrial deployment, remains largely unexplored in the research community:
\vspace{0pt}\par\noindent
\begin{tcolorbox}[
  colback=gray!8,
  colframe=black,
  boxrule=0.6pt,
  arc=3pt,
  left=7pt, right=7pt, top=1pt, bottom=1pt,
  before skip=7pt,
  after skip=7pt
]
\textit{\textbf{How can we train a tool-use agent for a specific application, with no pre-defined tasks, no verifiers, no simulator, and limited environment interaction?}}
\end{tcolorbox}

In this paper, we present \textbf{AgentBrew}, an offline training framework that learns effective tool-use policies from a single batch of raw interaction trajectories, without task verifiers or iterative rollouts. Our key insight is that every trajectory, regardless of whether it completes its original task, carries reusable signal: by inferring what each trajectory actually accomplished and scoring which actions genuinely contributed, we extract fine-grained supervision from the entire raw corpus without discarding any data.
\textbf{First}, the agent explores the target environment to discover its live state and proposes grounded task instructions, then executes these tasks to collect a raw trajectory corpus without any quality filtering (\S\ref{sec:experience collection}). \textbf{Second}, we recover fine-grained supervision from this noisy corpus through two complementary mechanisms (\S\ref{sec:filter}). For each trajectory, \emph{retrospective task inference} reconstructs an instruction that reflects what the trajectory actually achieved rather than what it originally intended, producing aligned instruction-trajectory pairs. 
\emph{PMI-Based credit assignment} then quantifies how much each action contributes to the inferred task. We define the total information a trajectory carries about its inferred instruction via pointwise mutual information (PMI), and leverage the chain rule of PMI to decompose this trajectory-level quantity into additive per-action credits, each measuring the marginal information a single action contributes. 
\textbf{Third}, these credits serve as sample weights for policy training, amplifying the gradient from informative actions while suppressing that from ineffective ones (\S\ref{sec:training}). Crucially, once the raw corpus is gathered, all subsequent stages proceed entirely offline with no further environment interaction, making the framework practical for real-world APIs where iterative exploration is infeasible.

We evaluate AgentBrew on three real-world MCP applications: GitHub, Notion, and PostgreSQL \citep{mcp-universe,mcpmark}. Without any external reward signals, human annotations, or iterative environment interaction, AgentBrew improves over the Qwen3-32B base model by +8.7 Acc and +9.7 Score on average, surpassing Qwen3-235B and outperforming rejection sampling by +5.9 Acc and +10.3 Score. These results demonstrate that fine-grained offline learning can extract useful training signal from raw trajectories that filtering-based approaches would discard entirely.

\section{Problem Formulation}
\label{sec:problem formulation}

We consider an LLM agent operating within a real-world application environment $\mathcal{E}$ (e.g., GitHub, Notion) that exposes a set of callable tools $\mathcal{F} = \{f_1, \ldots, f_K\}$, each defined by a natural language description and a typed parameter specification. Given a task instruction $I \in \mathcal{I}$, the agent interacts with $\mathcal{E}$ over discrete steps $t = 0, \ldots, T$. At each step, the agent observes the state $s_t = (I, a_0, o_0, \ldots, a_{t-1}, o_{t-1})$ and generates an action $a_t = (u_t, c_t)$, where $u_t$ is a chain-of-thought trace and $c_t$ is a tool call. The environment returns an observation $o_t = \mathcal{E}(c_t)$, yielding a complete trajectory $\tau = \{(s_t, a_t, o_t)\}_{t=0}^{T}$. 

Our goal is to improve the agent's policy $\pi_\theta$ for tool-use tasks in $\mathcal{E}$ under three constraints that reflect the practical reality of most real-world applications (\S\ref{sec: Introduction}): (1) no pre-defined tasks or verifiers are available; (2) the agent may collect a single trajectory corpus $\mathcal{D}_{\text{raw}} = \{(I_n, \tau_n)\}_{n=1}^{N}$ but further large-scale interaction is infeasible; and (3) trajectories carry no external reward signals. Each trajectory terminates when the agent issues a completion action or a maximum step limit is reached, but neither condition constitutes ground-truth verification of success. The objective is to extract a training signal from this unlabeled $\mathcal{D}_{\text{raw}}$ that improves $\pi_\theta$ on the distribution of user tasks in $\mathcal{E}$.

\section{Methodology}
\label{sec:method}
\begin{figure*}[t]
\centering
\includegraphics[width=\textwidth]{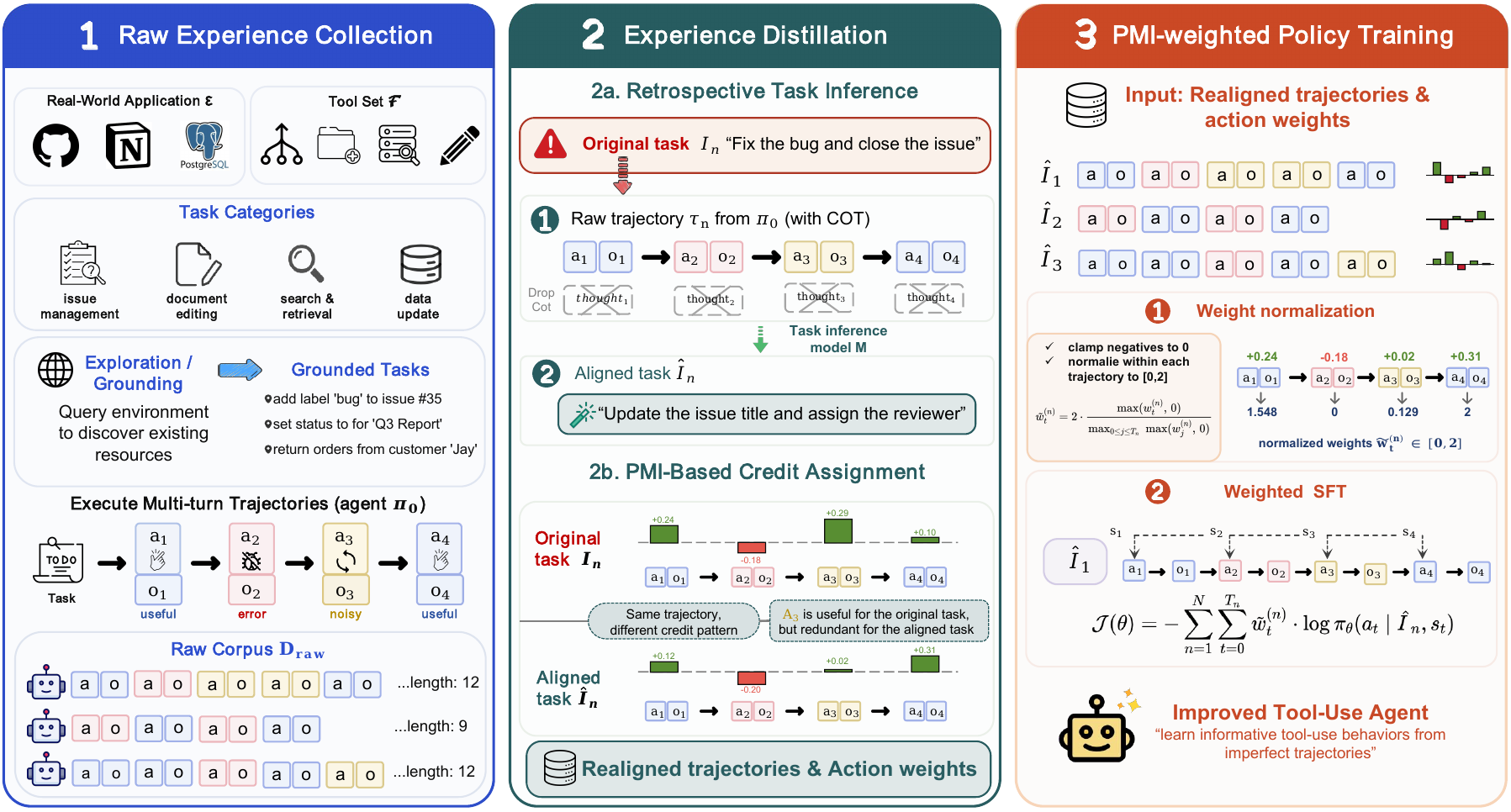}
\caption{Overview of AgentBrew. (1) The agent explores $\mathcal{E}$ to propose grounded tasks and collects raw trajectories. (2) Retrospective task inference recovers aligned instructions $\hat{I}_n$, and PMI-based credit assignment scores each action. (3) Normalized credits weight the policy training objective.}
\label{fig:overview}
\vspace{-3mm}
\end{figure*}

We propose \textbf{AgentBrew}, a three-stage offline framework illustrated in Figure~\ref{fig:overview}. Given a target environment $\mathcal{E}$, the agent first collects a raw trajectory corpus $\mathcal{D}_{\text{raw}}$ through grounded exploration (\S\ref{sec:experience collection}). The raw corpus then undergoes experience distillation (\S\ref{sec:filter}), which produces realigned instruction-trajectory pairs with per-action credit scores. Finally, these credits serve as sample weights for a maximum likelihood training objective (\S\ref{sec:training}). Once the initial corpus is gathered, the entire pipeline requires no further environment interaction.

\subsection{Raw Experience Collection}
\label{sec:experience collection}

\paragraph{Grounded Task Proposal.}
Prior approaches synthesize tasks directly from tool documentation \citep{toolllm, toucan}, but frequently produce tasks that reference nonexistent entities or invalid configurations because they lack access to the live state of $\mathcal{E}$. We instead ground task construction in the environment itself. The agent first generates a set of high-level task categories from the tool descriptions of $\mathcal{F}$ (see Appendix~\ref{app:task_proposal}), following the common practice of aligning synthetic training tasks with target capability profiles \citep{toucan, xlam, astra}. Each category specifies the capability profile to exercise, not the specific entities or data involved. Training and evaluation use fully isolated environment instances with no overlap in entities, data entries, or environment states. For each category, the agent conducts exploratory interactions with $\mathcal{E}$, invoking tools to retrieve concrete environmental information such as available resources and existing data entries, and then proposes task instructions grounded in the retrieved information. Generated tasks are validated to ensure all referenced schema entities exist in the target environment. This yields a task set $\mathcal{I} = \{I_n\}_{n=1}^{N}$ whose instructions reference entities and configurations that verifiably exist in $\mathcal{E}$.

\paragraph{Task Execution.}
For each proposed task $I_n$, the agent interacts with $\mathcal{E}$ over multiple turns following the process described in \S\ref{sec:problem formulation}, producing a trajectory $\tau_n = \{(s_t, a_t, o_t)\}_{t=0}^{T_n}$. The interaction terminates when the agent returns a final answer indicating task completion, or when a maximum turn limit $T_{\max}$ is reached. Trajectories that reach $T_{\max}$ are discarded as they typically reflect degenerate behaviors such as repeated failed API calls with no meaningful environmental interaction. All remaining naturally terminated trajectories are retained without quality filtering, yielding the raw corpus $\mathcal{D}_{\text{raw}} = \{(I_n, \tau_n)\}_{n=1}^{N}$.

\subsection{Experience Distillation}
\label{sec:filter}
The original instructions $\{I_n\}$ are necessary to drive trajectory collection, but most trajectories fail to complete $I_n$ \citep{mcp-universe,mcpmark,agentworld}, making the raw pairs $(I_n, \tau_n)$ unsuitable for direct training: the policy would learn to associate incorrect behaviors with the stated goal. We address this through two complementary steps. \emph{Retrospective task inference} (\S\ref{sec:realignment}) replaces each $I_n$ with a revised instruction $\hat{I}_n$ that reflects what the trajectory actually accomplished, producing aligned pairs for training. \emph{PMI-based credit assignment} (\S\ref{sec:credit}) then scores each action by how much it contributes to $\hat{I}_n$, so that informative actions receive stronger supervision while irrelevant or erroneous ones are suppressed.

\subsubsection{Retrospective Task Inference}
\label{sec:realignment}
To construct aligned pairs, we first discard all chain-of-thought traces $u_t$ from each trajectory, retaining only the tool calls and environment responses:
\begin{equation}
\bar{\tau}_n = \{(c_t, o_t)\}_{t=0}^{T_n}
\label{eq:mask}
\end{equation}
This removal is necessary because $u_t$ is generated conditioned on the original instruction $I_n$ and may contradict the trajectory's actual outcome: for instance, $u_t$ may claim a sub-goal is being achieved while $o_t$ indicates an API error or no state change. Retaining such traces would bias the inference toward intentions the trajectory did not fulfill. The resulting $\bar{\tau}_n$ is analogous to a user interaction log of application operations and system responses, stripped of subjective rationales.
 
Given $\bar{\tau}_n$, we prompt an LLM to infer a revised instruction $\hat{I}_n$ that is consistent with the trajectory's observable outcomes (prompt details in Appendix~\ref{app:prompt_inference}). The inference is guided by several key constraints: $\hat{I}_n$ must describe the outcome genuinely achieved in $\mathcal{E}$ rather than narrate the action sequence; claims not directly supported by successful operations in $\bar{\tau}_n$ are removed; and no entities or results beyond what is evidenced by the trajectory may be introduced. This produces realigned pairs $(\hat{I}_n, \bar{\tau}_n)$ for downstream supervision.

\subsubsection{PMI-Based Credit Assignment}
\label{sec:credit}
After retrospective task inference, we obtain aligned pairs $(\hat{I}_n, \bar{\tau}_n)$ that could in principle be used directly for supervised fine-tuning. However, not all actions in $\tau_n$ contribute equally to $\hat{I}_n$: some served the original intent of $I_n$ but are irrelevant to $\hat{I}_n$, and others are low-quality such as incorrect API calls and redundant queries. We therefore need a per-action credit that quantifies each action's contribution to $\hat{I}_n$. Our key idea is to use a reference language model $\pi_{\text{ref}}$ to measure how much information each action-observation pair provides about the inferred task $\hat{I}_n$. This informational measure correlates well with actual contribution because $\hat{I}_n$ is itself derived from the trajectory's observable outcomes (\S\ref{sec:realignment}): actions that produced the described effects leave distinctive traces in $o_t$ that strongly reduce uncertainty about the outcome-derived $\hat{I}_n$, receiving high credit, while failed or redundant actions provide little such evidence and are suppressed. We formalize this using pointwise mutual information (PMI).

\paragraph{Total information of a trajectory.} To assign per-action credits, we first define the total information that a trajectory carries about its inferred task, then decompose it across individual actions. Let $\pi_{\text{ref}}$ denote the reference language model. The PMI between the reduced trajectory $\bar{\tau}_n = \{(c_t, o_t)\}_{t=0}^{T_n}$ and the inferred instruction $\hat{I}_n$ is:
\begin{equation}
\text{PMI}(\hat{I}_n;\, \bar{\tau}_n) = \log \frac{\pi_{\text{ref}}(\hat{I}_n \mid \bar{\tau}_n)}{\pi_{\text{ref}}(\hat{I}_n)} = \underbrace{(-\log \pi_{\text{ref}}(\hat{I}_n))}_{\mathcal{L}_{\emptyset}} - \underbrace{(-\log \pi_{\text{ref}}(\hat{I}_n \mid \bar{\tau}_n))}_{\mathcal{L}_{T_n}} = \mathcal{L}_{\emptyset} - \mathcal{L}_{T_n}
\label{eq:pmi}
\end{equation}
where $\mathcal{L}_{\emptyset} = -\log \pi_{\text{ref}}(\hat{I}_n)$ is the negative log-likelihood (NLL) of $\hat{I}_n$ under the reference model without any context, and $\mathcal{L}_{T_n} = -\log \pi_{\text{ref}}(\hat{I}_n \mid \bar{\tau}_n)$ is the NLL after conditioning on the entire trajectory. We denote this total information as $\Delta \mathcal{L} = \mathcal{L}_{\emptyset} - \mathcal{L}_{T_n}$. The question is how to decompose $\Delta \mathcal{L}$ across the $T_n + 1$ individual actions.

\paragraph{Per-action credit via the chain rule of PMI.}
The chain rule of PMI decomposes the total information into a sum of conditional PMI terms, each measuring the marginal information contribution of a single action. Let $\mathcal{L}_t = -\log \pi_{\text{ref}}(\hat{I}_n \mid c_0, o_0, \ldots, c_t, o_t)$ denote the NLL after observing actions up to step $t$. The conditional PMI between $(c_t, o_t)$ and $\hat{I}_n$ given preceding actions $(c_{<t}, o_{<t})$ is:
\begin{equation}
w_t = \text{PMI}(\hat{I}_n;\, (c_t, o_t) \mid c_{<t}, o_{<t}) = \log \frac{\pi_{\text{ref}}(\hat{I}_n \mid c_{\leq t}, o_{\leq t})}{\pi_{\text{ref}}(\hat{I}_n \mid c_{<t}, o_{<t})} = \mathcal{L}_{t-1} - \mathcal{L}_t, \quad t = 1, \ldots, T_n
\label{eq:credit}
\end{equation}
with $w_0 = \mathcal{L}_{\emptyset} - \mathcal{L}_0$ for the first action. Each $w_t$ quantifies how much additional information $(c_t, o_t)$ provides about $\hat{I}_n$ beyond what is already known from the preceding actions. An action that is highly informative of $\hat{I}_n$ yields a large positive $w_t$, a redundant action yields $w_t \approx 0$, and an action whose observation misleads the model away from $\hat{I}_n$ yields $w_t < 0$.

By the chain rule of PMI, these per-action credits form an exact additive decomposition of the total information:
\begin{equation}
\sum_{t=0}^{T_n} w_t = (\mathcal{L}_{\emptyset} - \mathcal{L}_0) + \sum_{t=1}^{T_n}(\mathcal{L}_{t-1} - \mathcal{L}_t) = \mathcal{L}_{\emptyset} - \mathcal{L}_{T_n} = \Delta \mathcal{L}
\label{eq:decomposition}
\end{equation}
This guarantees that the per-action credits neither inflate nor discard any information: each unit of PMI between $\bar{\tau}_n$ and $\hat{I}_n$ is attributed to exactly one action.

\subsection{PMI-weighted Policy Training}
\label{sec:training}
The raw credits may be negative and vary in scale across trajectories. We clamp negative values to zero because any instrumental value of an uncertainty-increasing action is already captured by the subsequent step that resolves that uncertainty, as guaranteed by the additive decomposition in Eq.~(\ref{eq:decomposition}). We then normalize within each trajectory to $[0, 2]$:
\begin{equation}
\tilde{w}_t^{(n)} = 2 \cdot \frac{\max(w_t^{(n)},\, 0)}{\max_{0 \leq j \leq T_n} \, \max(w_j^{(n)},\, 0)}
\label{eq:normalize}
\end{equation}
The policy is then optimized by minimizing the weighted negative log-likelihood, following the sample-weighted SFT paradigm used in prior work \citep{rft, rest}:
\begin{equation}
\mathcal{J}(\theta) = -\sum_{n=1}^{N} \sum_{t=0}^{T_n} \tilde{w}_t^{(n)} \cdot \log \pi_\theta(a_t \mid \hat{I}_n, s_t)
\label{eq:objective}
\end{equation}
where $s_t = (\hat{I}_n,\, a_0, o_0, \ldots, a_{t-1}, o_{t-1})$. Although credits are computed over $(c_t, o_t)$, the loss is applied to the full action $a_t = (u_t, c_t)$ including chain-of-thought tokens. The credit weighting mitigates potential misalignment between $u_t$ and $\hat{I}_n$: low-contribution actions receive near-zero weight, suppressing gradient from both their tool calls and associated reasoning, while high-credit actions tend to have reasoning consistent with their executed behavior. Overall, informative actions receive amplified gradient signal while uninformative ones are effectively masked. We present pseudo code and additional details in Appendix~\ref{app:algorithm} \&~\ref{app:pmi-credit-details}.

\newcommand{\ticon}[1]{\raisebox{-0.15em}{\includegraphics[height=1em]{Fingure/icons/#1}}\,}

\newcommand{\gain}[1]{{\scriptsize\textcolor{green!50!black}{(+#1)}}}
\newcommand{\loss}[1]{{\scriptsize\textcolor{red}{(#1)}}}

\begin{table*}[t]
\centering
\vspace{-1mm}
\caption{Main results across three real-world applications. We report \textbf{Acc} (binary task completion rate) and \textbf{Score} (average per-task score, 0--1). Best results per column are \textbf{bolded}. {\scriptsize\textcolor{green!50!black}{Green}} and {\scriptsize\textcolor{red}{red}} deltas indicate changes relative to Qwen3-32B.}
\vspace{4pt}
\small
\setlength{\tabcolsep}{7.5pt}
\renewcommand{\arraystretch}{1.15}
\begin{tabular}{l cc cc cc cc}
\toprule
\multirow{2.4}{*}{\textbf{Method}} 
& \multicolumn{2}{c}{\textbf{GitHub}} 
& \multicolumn{2}{c}{\textbf{Notion}} 
& \multicolumn{2}{c}{\textbf{Postgres}}
& \multicolumn{2}{c}{\textbf{Average}} \\
\cmidrule(lr){2-3} \cmidrule(lr){4-5} \cmidrule(lr){6-7} \cmidrule(lr){8-9}
& Acc. & Score & Acc. & Score & Acc. & Score & Acc. & Score \\
\midrule
\rowcolor{gray!15}
\multicolumn{9}{c}{\textbf{Frontier Proprietary Models}} \\[2pt]
\ticon{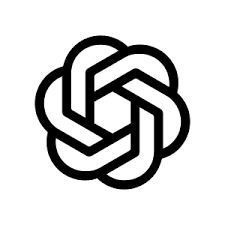} GPT-5              & 64.2 & 85.4 & 42.9 & 83.4 & 42.9 & 61.7 & 50.0 & 76.8 \\
\ticon{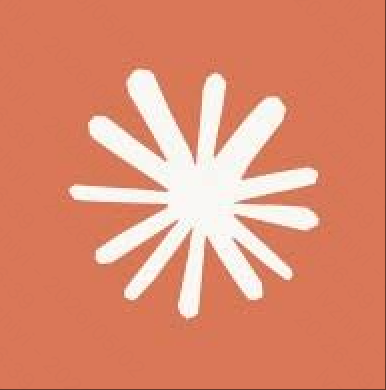} Claude Sonnet-4.5  & 57.1 & 76.9 & 39.3 & 84.6 & 38.1 & 53.3 & 44.8 & 71.6 \\
\ticon{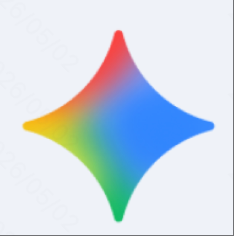} Gemini-3 Pro       & 60.7 & 81.2 & 53.6 & 86.2 & 57.1 & 75.1 & 57.1 & 80.8\relax \\[3pt]
\rowcolor{gray!15}
\multicolumn{9}{c}{\textbf{Open-Source Foundation Models}} \\[2pt]
\ticon{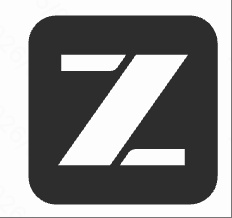} GLM-5-744B            & 42.9 & 74.3 & 32.1 & 77.3 & 23.8 & 59.4 & 32.9 & 70.3 \\
\ticon{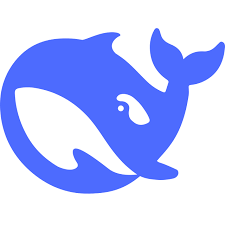} DeepSeek-V3.2-685B & 17.8 & 63.1 & 39.2 & 74.3 & 33.3 & 64.1 & 30.1 & 67.2 \\
\ticon{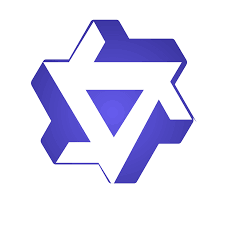} Qwen3-235B           & 17.9 & 63.0 & 10.7 & 65.7 & 19.0 & 34.3 & 15.9 & 54.3 \\
\ticon{gpt} GPT-OSS-120B       & 3.6 & 37.8 & 3.6 & 54.6 & 14.3 & 24.1 & 7.1 & 38.8 \\
\ticon{qwen} Qwen3-32B            & 10.7 & 49.7 & 3.6 & 60.1 & 14.3 & 37.2 & 9.5 & 49.0\relax \\[3pt]
\rowcolor{gray!15}
\multicolumn{9}{c}{\textbf{Training Methods (32B)}} \\[2pt]
\ticon{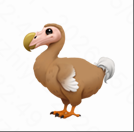} TOUCAN              & 17.8 & 56.4 & 3.6 & 49.2 & 14.3 & 28.8 & 11.9 \gain{2.4} & 44.8 \loss{-4.2} \\
\ticon{qwen} Vanilla SFT          & 14.2 & 53.9 & 7.1 & 59.8 & 9.5 & 30.7 & 10.3 \gain{0.8} & 48.1 \loss{-0.9} \\
\ticon{qwen} Rejection Sampling   & 17.8 & 53.9 & 0 & 53.2 & 19.0 & 38.0 & 12.3 \gain{2.8} & 48.4 \loss{-0.6} \\
\hdashline\noalign{\vskip 3pt}
\rowcolor{blue!5}
\ticon{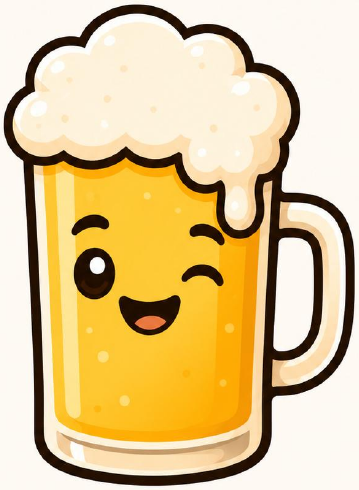} \textbf{AgentBrew} & \textbf{21.4} & \textbf{63.5} & \textbf{14.2} & \textbf{70.1} & \textbf{23.8} & \textbf{41.8} & \textbf{18.2} \gain{8.7} & \textbf{58.7} \gain{9.7} \\
\bottomrule
\end{tabular}
\label{tab:main_results}
\end{table*}

\section{Experiments}
\label{sec:experiments}
We evaluate AgentBrew across three real-world applications, examining overall performance (\S\ref{sec:main_results}), the effectiveness of PMI-based credit assignment and retrospective task inference (\S\ref{sec:analysis_credit}--\ref{sec:analysis_hindsight}), robustness to noisy data (\S\ref{sec:analysis_robust}), and transferability across model sizes (\S\ref{sec:analysis_transfer}).

\subsection{Experimental Settings}
\label{sec:exp_setup}
 
\paragraph{Setup.}
We evaluate on three real-world MCP applications \citep{mcp-universe, mcpmark}: \textbf{GitHub} (code collaboration), \textbf{Notion} (knowledge management), and \textbf{PostgreSQL} (relational database). We compare against frontier proprietary models (GPT-5, Claude Sonnet-4.5, Gemini-3 Pro), open-source foundation models (GLM-5-744B, DeepSeek-V3.2-685B, Qwen3-235B, GPT-OSS-120B, Qwen3-32B), and training-based methods built on Qwen3-32B: TOUCAN \citep{toucan}, Vanilla SFT on our collected trajectories, and Rejection Sampling \citep{rft} which retains only LLM-judge-approved trajectories for SFT. The evaluation tasks and ground-truth verifiers are provided by the mcp-universe and mcpmark benchmarks \citep{mcp-universe, mcpmark}; that these verifiers are used solely for evaluation; they are not available during any stage of the AgentBrew pipeline. To prevent data leakage, training and evaluation are isolated at two levels: training tasks are generated from capability descriptions, and the training corpus is collected from environment instances that share no entities, data entries, or states with the benchmark evaluation environments (details in Appendix \ref{app:implementation}).

\paragraph{Data collection and implementation.}
We use Qwen3-32B \citep{qwen3} as the unified model for the entire AgentBrew pipeline, serving as the base policy, the reference model $\pi_{\text{ref}}$ for PMI credit computation, and the backbone for retrospective task inference. The raw corpus $\mathcal{D}_{\text{raw}}$ contains approximately 2,500  trajectories for GitHub, 2,500 for Notion, and 3,500 for PostgreSQL. Since real-world interactions often exceed 100K tokens, we adopt a memory mechanism that retains the three most recent steps in full and compresses earlier history into LLM-generated summaries \citep{memGPT,recurrentgpt}. This mechanism is applied uniformly across all methods and models, ensuring fair comparison. Since all prefixes of a trajectory share a common prompt, PMI credit computation benefits from KV-cache reuse and can be parallelized across trajectories; computing credits for the entire corpus of approximately 8,500 trajectories completes in roughly 9 hours on 4 H100 GPUs. Further details on data collection and implementation are provided in Appendix~\ref{app:implementation}, and case studies are presented in Appendix~\ref{app:case_studies}.

\subsection{Main Results}
\label{sec:main_results}
Table~\ref{tab:main_results} summarizes performance across three real-world applications. AgentBrew targets data efficiency rather than model scaling: it extracts stronger signal from limited, unlabeled interactions and is therefore complementary to advances in base model capability. Built on Qwen3-32B, AgentBrew achieves an average Acc of 18.2 and Score of 58.7, surpassing its own 235B variant (+2.3 Acc, +4.4 Score), demonstrating that targeted offline adaptation on a single environment can be more effective than general-purpose scaling alone. Absolute accuracy across all methods remains modest, reflecting the genuine difficulty of these real-world benchmarks, where even frontier proprietary models such as Gemini-3 Pro (57.1 / 80.8) and GPT-5 (50.0 / 76.8) have ample room for improvement.

Among training-based methods, AgentBrew consistently outperforms all baselines across the three environments. Compared to Vanilla SFT and Rejection Sampling, AgentBrew achieves improvements of +7.9 Acc / +10.6 Score and +5.9 Acc / +10.3 Score, respectively, showing that retrospectively realigning task instructions and extracting fine-grained per-action credit from noisy trajectories is more effective than either uniform supervision or trajectory-level quality filtering. The advantage is most pronounced on Notion, where Rejection Sampling yields an Acc of 0: the LLM judge fails to reliably assess task completion in complex stateful API interactions, causing the filter to reject nearly all trajectories and leaving insufficient data for effective training. AgentBrew achieves 14.2 Acc on the same corpus by bypassing verification entirely and instead recovering informative actions within imperfect trajectories. AgentBrew also outperforms TOUCAN (+6.3 Acc, +13.9 Score), which is trained on 1.5M trajectories synthesized across broad MCP environments, further confirming that a small amount of environment-specific data with fine-grained credit assignment is more effective than large-scale cross-environment trajectory synthesis.

\begin{figure*}[t]
\centering
\begin{minipage}[b]{0.24\textwidth}
    \centering
    \includegraphics[width=\textwidth]{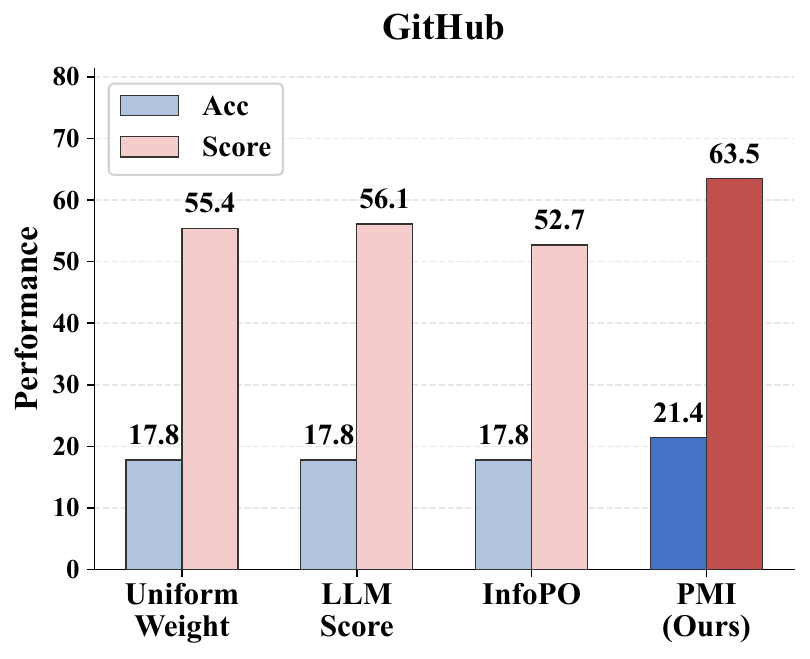}
    \vspace{-4pt}
    \centerline{\small (a)}
\end{minipage}
\hfill
\begin{minipage}[b]{0.24\textwidth}
    \centering
    \includegraphics[width=\textwidth]{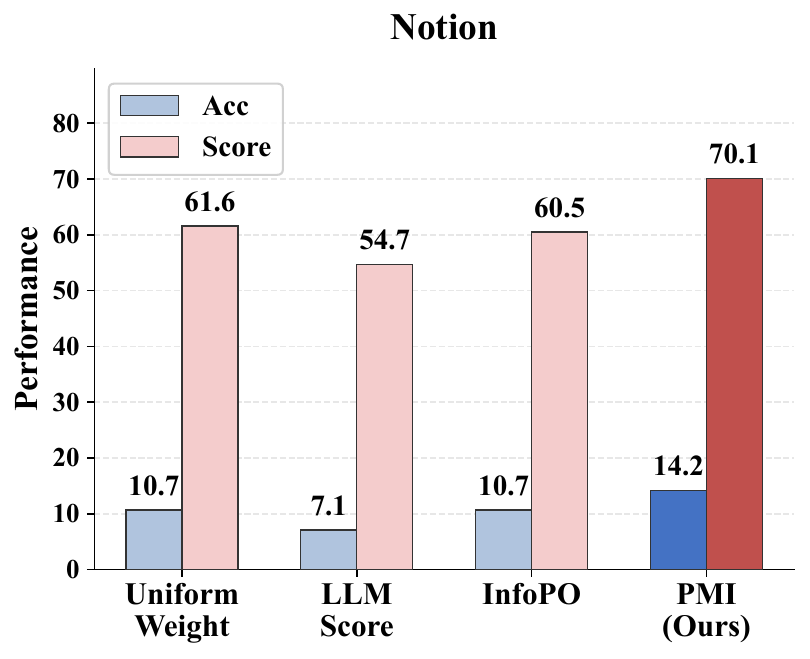}
    \vspace{-4pt}
    \centerline{\small (b)}
\end{minipage}
\hfill
\begin{minipage}[b]{0.24\textwidth}
    \centering
    \includegraphics[width=\textwidth]{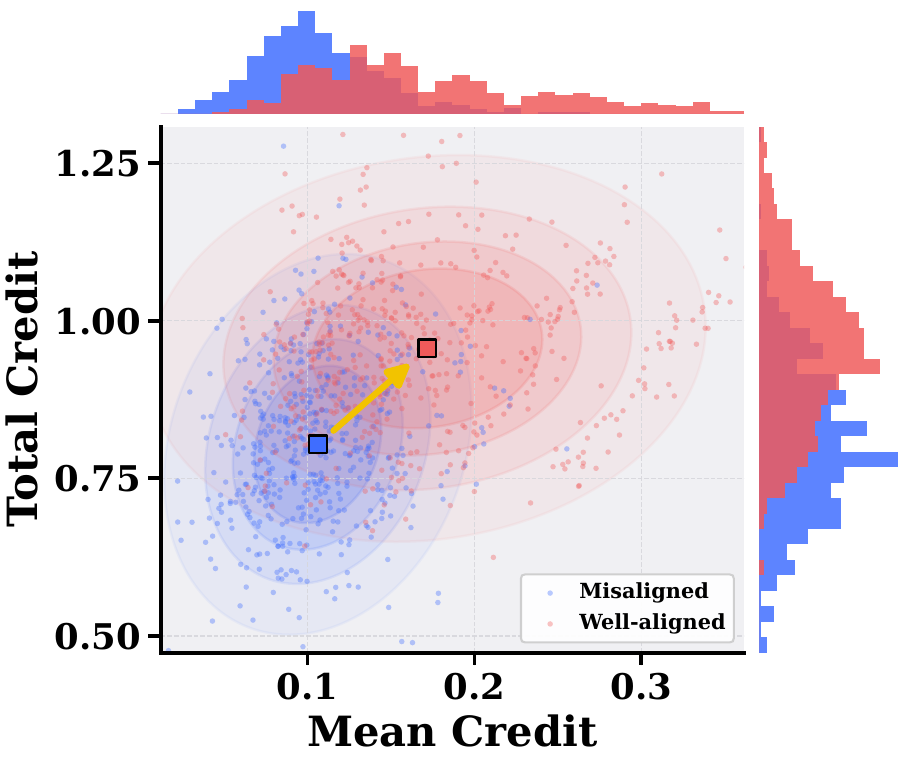}
    \vspace{-4pt}
    \centerline{\small (c)}
\end{minipage}
\hfill
\begin{minipage}[b]{0.24\textwidth}
    \centering
    \includegraphics[width=\textwidth]{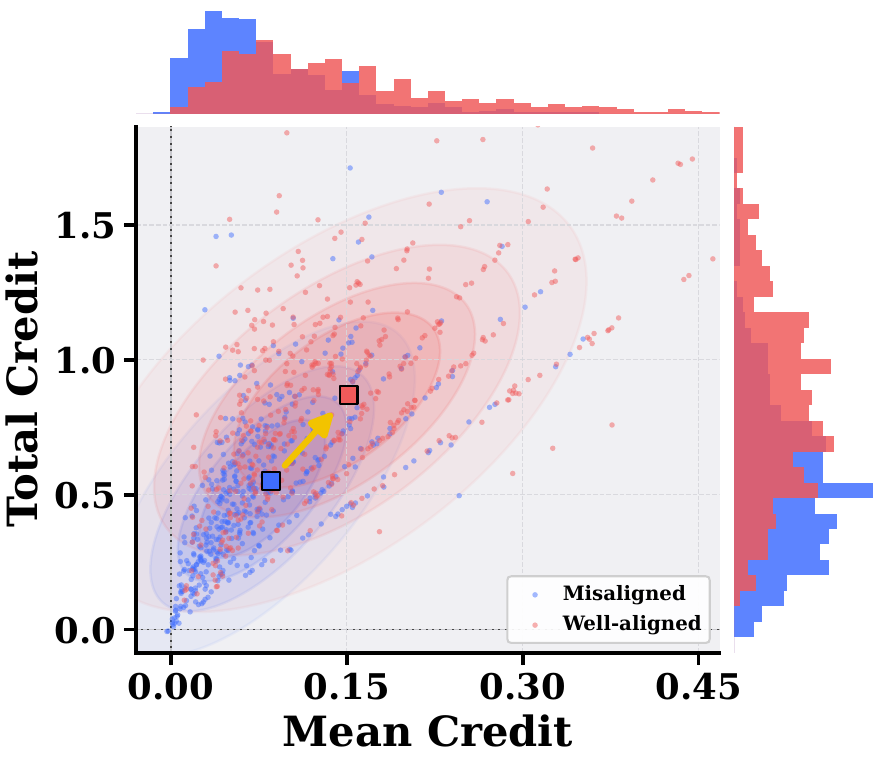}
    \vspace{-4pt}
    \centerline{\small (d)}
\end{minipage}
\caption{\textbf{Effect of credit assignment.} (a-b) Performance comparison of different credit assignment strategies on GitHub and Notion. (c-d) Distribution of mean credit and total credit for well-aligned (red) and misaligned (blue) trajectories on GitHub (c) and Notion (d).}
\label{fig:credit}
\vspace{-5mm}
\end{figure*}
\subsection{Effect of PMI-Based Credit Assignment}
\label{sec:analysis_credit}

We examine how PMI-Based credit assignment (\S\ref{sec:credit}) identifies informative actions within imperfect trajectories. All methods are trained on the same realigned pairs $\{(\hat{I}_n, \tau_n)\}$ from retrospective task inference (\S\ref{sec:realignment}); only the per-action weighting strategy differs. We compare our PMI-based credit against three alternatives: \emph{Uniform}, which assigns equal weight to all actions; \emph{LLM Score}, which prompts $\pi_{\text{ref}}$ to score each of its own actions; and \emph{InfoPO} \citep{infopo}, which computes credit based on the information gain advantage over future actions. As shown in Figure~\ref{fig:credit}(a-b), PMI-based credit consistently outperforms all alternatives on both GitHub and Notion. Notably, LLM Score drops to 7.1 / 54.7 on Notion, underperforming even Uniform (10.7 / 61.6), suggesting that LLM self-evaluation is unreliable for action-level quality assessment in complex stateful tool-use settings.

Beyond downstream performance, PMI-based credit provides a natural safeguard against imperfect task inference: when $\hat{I}_n$ is poorly aligned with the trajectory, actions provide little information about it, yielding uniformly low credits that are clamped to near-zero by Eq.~(\ref{eq:normalize}) and ensuring that the trajectory contributes minimal gradient signal during training. We verify this by using GPT-5 to classify each $(\hat{I}_n, \bar{\tau}_n)$ pair as well-aligned or misaligned. As shown in Figure~\ref{fig:credit}(c-d), misaligned trajectories (blue) cluster at significantly lower mean and total credit compared to well-aligned ones (red) on both GitHub and Notion, confirming that PMI-based credit automatically suppresses noisy task inference without requiring any explicit quality filter.

\subsection{Effect of Retrospective Task Inference}
\label{sec:analysis_hindsight}
 
\begin{wraptable}{r}{0.4\textwidth}
  \vspace{-12pt}
  \centering
  \small
  \caption{Effect of retrospective task inference on Notion.}
  \label{tab:hindsight}
  \begin{tabular}{lcc}
  \toprule
  \textbf{Task Instruction} & \textbf{Acc.} & \textbf{Score} \\
  \midrule
  Original $I_n$     & 7.1 & 65.6 \\
  Inferred $\hat{I}_n$ & \textbf{14.2} & \textbf{70.1} \\
  \bottomrule
  \end{tabular}
  \vspace{-10pt}
\end{wraptable}
 
We verify the necessity of retrospective task inference (\S\ref{sec:realignment}) by comparing two training configurations on Notion: one using the original task instructions $\{I_n\}$ and the other using the inferred instructions $\{\hat{I}_n\}$, both trained with PMI-based credit assignment. As shown in Table~\ref{tab:hindsight}, training with $\hat{I}_n$ achieves an Acc of 14.2 and Score of 70.1, compared to 7.1 / 65.6 with the original $I_n$. The degradation under $I_n$ arises because most trajectories fail to complete their original instruction, causing the policy to associate incomplete or erroneous action sequences with the stated goal (see Appendix~\ref{app:case_studies} for qualitative examples). Although PMI-based credit assignment can down-weight misaligned trajectories (\S\ref{sec:analysis_credit}), credit is computed as the information gain of each action with respect to whatever instruction is provided; when the instruction itself is misaligned, even useful actions receive distorted credits measured against the wrong reference task. Retrospective task inference addresses this at the source by replacing $I_n$ with $\hat{I}_n$ that reflects the trajectory's actual outcome, ensuring that credit computation operates on coherent instruction-trajectory pairs. The two mechanisms are thus complementary: task inference corrects \emph{what} the trajectory is trained to achieve, while credit assignment determines \emph{how much} each action contributes to that corrected goal.

\subsection{Robustness to Noisy Trajectories}
\label{sec:analysis_robust}
\begin{wrapfigure}{r}{0.4\textwidth}
  \vspace{-12pt}
  \centering
  \includegraphics[width=0.38\textwidth]{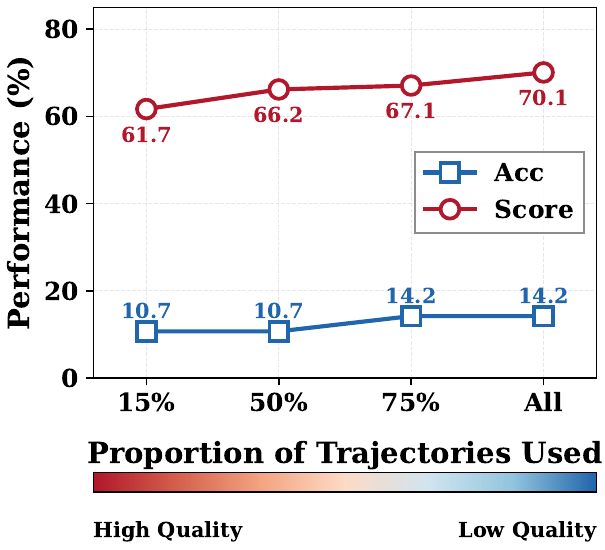}
  \caption{Performance on Notion when progressively adding lower-quality data.}
  \label{fig:robustness}
  \vspace{-10pt}
\end{wrapfigure}
As an offline method, AgentBrew can in principle be applied to trajectories from diverse sources, including agent self-exploration and human usage logs, where data quality is inherently uncontrolled. Robustness to such quality variance is therefore a practical requirement. To evaluate this, we use GPT-5 to score each raw trajectory pair $(I_n, \tau_n)$ based on instruction-trajectory alignment and rank all trajectories by quality in descending order. We then train AgentBrew on progressively larger subsets: the top 15\%, top 50\%, top 75\%, and the full corpus, where each expansion introduces trajectories of strictly lower quality than those already included.
 
As shown in Figure~\ref{fig:robustness}, both metrics improve monotonically as lower-quality data is added, with Acc increasing from 10.7 to 14.2 and Score from 61.7 to 70.1. Even adding the lowest-quality 25\% of trajectories yields a further Score gain (67.1 $\rightarrow$ 70.1). This does not imply that lower-quality trajectories are inherently more valuable; rather, each additional trajectory, regardless of its overall completion quality, still contains individual actions that are informative for the inferred task. AgentBrew's credit mechanism extracts these useful portions while suppressing the rest, so the net effect of including more data is consistently positive. This allows practitioners to include trajectories from diverse and uncontrolled sources without degrading performance.

\subsection{Offline Transferability Across Models}
\label{sec:analysis_transfer}
To test whether AgentBrew's distilled signal transfers across model sizes, we apply the data distilled by Qwen3-32B to train a 14B model via the same weighted training objective. As shown in Table~\ref{tab:transfer}, AgentBrew-14B achieves 20.2 average Acc / 55.3 Score, substantially outperforming the Qwen3-14B base model (+11.5 Acc / +12.5 Score), and on Notion nearly matches the 32B AgentBrew model (14.2 / 69.8 vs 14.2 / 70.1). Toucan-14B and AWM-14B also leverage larger models in their pipelines for trajectory synthesis and environment construction, yet show limited improvement over the base model, indicating that the transferability stems from the quality of AgentBrew's distilled signal rather than the capacity gap alone. This confirms that a single round of offline distillation can be reused to train smaller models, amortizing the cost of real-world API access.

\begin{table}[t]
\centering
\caption{Offline transferability on 14B models. AgentBrew distills data collected by Qwen3-32B into a 14B model via the offline pipeline, without additional environment interaction.}
\label{tab:transfer}
\resizebox{\linewidth}{!}{
\begin{tabular}{llcccccccc}
\toprule
\textbf{Method} & \textbf{Training Paradigm}
& \multicolumn{2}{c}{\textbf{GitHub}}
& \multicolumn{2}{c}{\textbf{Notion}}
& \multicolumn{2}{c}{\textbf{Postgres}}
& \multicolumn{2}{c}{\textbf{Avg.}} \\
\cmidrule(lr){3-4}
\cmidrule(lr){5-6}
\cmidrule(lr){7-8}
\cmidrule(lr){9-10}
& & Acc. & Score
  & Acc. & Score
  & Acc. & Score
  & Acc. & Score \\
\midrule
Qwen3-14B & No additional training
& 7.1 & 48.2 & 0 & 48.8 & 19.0 & 31.3 & 8.7 & 42.8 \\
Toucan-14B & SFT on 1.5M synthetic trajs.
& 0 & 45.2 & 0 & 42.5 & 0 & 11.4 & 0.0 & 33.3 \\
AWM-14B & Online RL on synthetic envs.
& 7.1 & 45.4 & 0 & 44.5 & 23.8 & 38.0 & 10.3 & 42.6 \\
\midrule
\textbf{AgentBrew-14B} & \textbf{Offline transfer}
& \textbf{17.9} & \textbf{56.3}
& \textbf{14.2} & \textbf{69.8}
& \textbf{28.6} & \textbf{40.0}
& \textbf{20.2} & \textbf{55.3} \\
\bottomrule
\end{tabular}
}
\vspace{-3mm}
\end{table}

\section{Related Work}
\label{sec:related_work}

\paragraph{Training Data for Tool-Use Agents.}
A central challenge in training tool-use agents is obtaining high-quality (task, trajectory) pairs at scale. The prevailing forward approach first defines tasks, then collects trajectories by executing them with strong models and retaining only successful completions \citep{toolllm, xlam, apigen-mt, toucan, astra, simia}. An alternative reverse approach derives tasks from existing interactions: OS-Genesis \citep{os-genesis} infers tasks from GUI state transitions, AgentTrek \citep{agenttrek} uses web tutorials to guide trajectory synthesis, and Learn-by-interact \citep{learn_by_interact} synthesizes trajectories from documentation and constructs instructions via backward abstraction of the interaction histories. Closer to our work, hindsight relabeling methods recover training signal from failed trajectories by replacing the original instruction with one the trajectory actually fulfills. This idea originates from HER \citep{her} in classic RL; HSL \citep{hsl} adapts it to LLM agents using environment rewards, and AgentHER \citep{agenther} introduces failure classification with confidence gating. Across these lines of work, data quality is ensured through external signals such as execution verification, environment rewards, or LLM-based filtering. AgentBrew requires none of these, instead deferring quality control entirely to an information-theoretic credit assignment mechanism that operates on the trajectories themselves.

\paragraph{Policy Optimization for LLM Agents.}
On-policy reinforcement learning is widely used to train LLM agents in interactive environments, with methods such as GRPO \citep{grpo}, DAPO \citep{dapo}, WebRL \citep{webrl}, and DigiRL \citep{digirl} relying on iterative rollouts and environment rewards. A central challenge is credit assignment, since trajectory-level rewards provide sparse supervision and cannot distinguish informative actions from redundant ones. To obtain finer-grained signals, IGPO \citep{igpo} uses information gain on a ground-truth answer as turn-level reward, GiGPO \citep{gigpo} builds step-level groups from anchor states, iStar \citep{istar} infers step rewards from trajectory preferences, and AgentPRM \citep{agentprm} trains a process reward model via temporal-difference estimation. These methods remain on-policy and require repeated interaction with reward feedback. Offline methods reduce rollout cost: DPO \citep{dpo}, KTO \citep{kto}, and RPO \citep{rpo} learn from preference pairs, while Digi-Q \citep{digiq} and OREO \citep{ored} learn value or Q-functions from offline trajectories. Yet they still depend on rewards, preferences, or learned value functions. In contrast, AgentBrew is fully offline and derives per-action training weights directly from the relationship between trajectories and their inferred tasks, without external reward signals or preference annotations.

\section{Conclusion, Limitation, and Future Work}
\label{sec:conclusion}
We presented AgentBrew, an offline framework that trains tool-use agents from a single batch of raw interactions without task verifiers or on-policy rollouts. By retrospectively inferring aligned instructions and decomposing per-action credit, AgentBrew extracts training signal from noisy trajectories that filtering-based methods would discard, enabling a 32B model to surpass Qwen3-235B across three real-world applications. Current limitations include evaluation on only three MCP applications, potential degradation of task inference on extremely long or chaotic trajectories, and the reliance on PMI as an informational proxy for action quality, which, despite structural alignment with actual contribution through the construction of $\hat{I}_n$ (\S\ref{sec:credit}), may not capture all causal effects in highly entangled action sequences.

A promising future direction is applying AgentBrew to real user interaction logs in production applications, which are naturally unlabeled and noisy. Retrospective task inference could recover user intent directly from operation sequences, and credit assignment could identify key actions within each session, enabling agents to learn from authentic human demonstrations at scale.

\bibliographystyle{plainnat}
\bibliography{references}

\begin{thebibliography}{49}
\providecommand{\natexlab}[1]{#1}
\providecommand{\url}[1]{\texttt{#1}}
\expandafter\ifx\csname urlstyle\endcsname\relax
  \providecommand{\doi}[1]{doi: #1}\else
  \providecommand{\doi}{doi: \begingroup \urlstyle{rm}\Url}\fi

\bibitem[Andrychowicz et~al.(2017)Andrychowicz, Wolski, Ray, Schneider, Fong, Welinder, McGrew, Tobin, Abbeel, and Zaremba]{her}
Marcin Andrychowicz, Filip Wolski, Alex Ray, Jonas Schneider, Rachel Fong, Peter Welinder, Bob McGrew, Josh Tobin, Pieter Abbeel, and Wojciech Zaremba.
\newblock Hindsight experience replay.
\newblock \emph{Advances in neural information processing systems}, 30, 2017.

\bibitem[{Anthropic}(2024)]{mcp}
{Anthropic}.
\newblock Introducing the model context protocol.
\newblock \url{https://www.anthropic.com/news/model-context-protocol}, November 2024.

\bibitem[Bai et~al.(2024)Bai, Zhou, Cemri, Pan, Suhr, Levine, and Kumar]{digirl}
Hao Bai, Yifei Zhou, Mert Cemri, Jiayi Pan, Alane Suhr, Sergey Levine, and Aviral Kumar.
\newblock Digirl: Training in-the-wild device-control agents with autonomous reinforcement learning.
\newblock \emph{Advances in Neural Information Processing Systems}, 37:\penalty0 12461--12495, 2024.

\bibitem[Bai et~al.(2025)Bai, Zhou, Li, Levine, and Kumar]{digiq}
Hao Bai, Yifei Zhou, Li~Erran Li, Sergey Levine, and Aviral Kumar.
\newblock Digi-q: Learning q-value functions for training device-control agents.
\newblock \emph{arXiv preprint arXiv:2502.15760}, 2025.

\bibitem[Ding(2026)]{agenther}
Liang Ding.
\newblock Agenther: Hindsight experience replay for llm agent trajectory relabeling.
\newblock \emph{arXiv preprint arXiv:2603.21357}, 2026.

\bibitem[Dong et~al.(2026)Dong, Lu, Huang, Zhong, Liu, Huang, Li, Zhao, Song, Li, Jin, Zhu, Wang, Lei, Luo, Chen, Chen, Feng, Wen, and Dou]{agentworld}
Guanting Dong, Junting Lu, Junjie Huang, Wanjun Zhong, Longxiang Liu, Shijue Huang, Zhenyu Li, Yang Zhao, Xiaoshuai Song, Xiaoxi Li, Jiajie Jin, Yutao Zhu, Hanbin Wang, Fangyu Lei, Qinyu Luo, Mingyang Chen, Zehui Chen, Jiazhan Feng, Ji-Rong Wen, and Zhicheng Dou.
\newblock Agent-world: Scaling real-world environment synthesis for evolving general agent intelligence.
\newblock \emph{arXiv preprint}, 2026.

\bibitem[Drouin et~al.(2024)Drouin, Gasse, Caccia, Laradji, Del~Verme, Marty, Boisvert, Thakkar, Cappart, Vazquez, et~al.]{workarena}
Alexandre Drouin, Maxime Gasse, Massimo Caccia, Issam~H Laradji, Manuel Del~Verme, Tom Marty, L{\'e}o Boisvert, Megh Thakkar, Quentin Cappart, David Vazquez, et~al.
\newblock Workarena: How capable are web agents at solving common knowledge work tasks?
\newblock \emph{arXiv preprint arXiv:2403.07718}, 2024.

\bibitem[Ethayarajh et~al.(2024)Ethayarajh, Xu, Muennighoff, Jurafsky, and Kiela]{kto}
Kawin Ethayarajh, Winnie Xu, Niklas Muennighoff, Dan Jurafsky, and Douwe Kiela.
\newblock Kto: Model alignment as prospect theoretic optimization.
\newblock \emph{arXiv preprint arXiv:2402.01306}, 2024.

\bibitem[Feng et~al.(2025)Feng, Xue, Liu, and An]{gigpo}
Lang Feng, Zhenghai Xue, Tingcong Liu, and Bo~An.
\newblock Group-in-group policy optimization for llm agent training.
\newblock \emph{arXiv preprint arXiv:2505.10978}, 2025.

\bibitem[{GitHub}(2025)]{github_mcp_server}
{GitHub}.
\newblock Github mcp server.
\newblock \url{https://github.com/github/github-mcp-server}, 2025.
\newblock GitHub repository. Accessed: 2026-04-17.

\bibitem[Gulcehre et~al.(2023)Gulcehre, Paine, Srinivasan, Konyushkova, Weerts, Sharma, Siddhant, Ahern, Wang, Gu, et~al.]{rest}
Caglar Gulcehre, Tom~Le Paine, Srivatsan Srinivasan, Ksenia Konyushkova, Lotte Weerts, Abhishek Sharma, Aditya Siddhant, Alex Ahern, Miaosen Wang, Chenjie Gu, et~al.
\newblock Reinforced self-training (rest) for language modeling.
\newblock \emph{arXiv preprint arXiv:2308.08998}, 2023.

\bibitem[Guo et~al.(2024)Guo, Cheng, Wang, Liang, Qin, Li, Liu, Sun, and Liu]{stabletoolbench}
Zhicheng Guo, Sijie Cheng, Hao Wang, Shihao Liang, Yujia Qin, Peng Li, Zhiyuan Liu, Maosong Sun, and Yang Liu.
\newblock Stabletoolbench: Towards stable large-scale benchmarking on tool learning of large language models.
\newblock In \emph{Findings of the Association for Computational Linguistics: ACL 2024}, pages 11143--11156, 2024.

\bibitem[He et~al.(2026)He, Feng, Wei, Cheng, Feng, and An]{hehierarchy}
Shuo He, Lang Feng, Qi~Wei, Xin Cheng, Lei Feng, and Bo~An.
\newblock Hierarchy-of-groups policy optimization for long-horizon agentic tasks.
\newblock \emph{arXiv preprint arXiv:2602.22817}, 2026.

\bibitem[Kong et~al.(2026)Kong, Zhang, Deng, Wu, Luo, and Liu]{infopo}
Fanqi Kong, Jiayi Zhang, Mingyi Deng, Chenglin Wu, Yuyu Luo, and Bang Liu.
\newblock Infopo: Information-driven policy optimization for user-centric agents.
\newblock \emph{arXiv preprint arXiv:2603.00656}, 2026.

\bibitem[Lei et~al.(2024)Lei, Chen, Ye, Cao, Shin, Su, Suo, Gao, Hu, Yin, et~al.]{spider}
Fangyu Lei, Jixuan Chen, Yuxiao Ye, Ruisheng Cao, Dongchan Shin, Hongjin Su, Zhaoqing Suo, Hongcheng Gao, Wenjing Hu, Pengcheng Yin, et~al.
\newblock Spider 2.0: Evaluating language models on real-world enterprise text-to-sql workflows.
\newblock \emph{arXiv preprint arXiv:2411.07763}, 2024.

\bibitem[Li et~al.(2023)Li, Hui, Qu, Yang, Li, Li, Wang, Qin, Geng, Huo, et~al.]{bird}
Jinyang Li, Binyuan Hui, Ge~Qu, Jiaxi Yang, Binhua Li, Bowen Li, Bailin Wang, Bowen Qin, Ruiying Geng, Nan Huo, et~al.
\newblock Can llm already serve as a database interface? a big bench for large-scale database grounded text-to-sqls.
\newblock \emph{Advances in Neural Information Processing Systems}, 36:\penalty0 42330--42357, 2023.

\bibitem[Li et~al.(2025)Li, Inan, Yue, Chen, Wutschitz, Kulkarni, Poovendran, Sim, and Rajmohan]{simia}
Yuetai Li, Huseyin~A Inan, Xiang Yue, Wei-Ning Chen, Lukas Wutschitz, Janardhan Kulkarni, Radha Poovendran, Robert Sim, and Saravan Rajmohan.
\newblock Simulating environments with reasoning models for agent training.
\newblock \emph{arXiv preprint arXiv:2511.01824}, 2025.

\bibitem[Li et~al.(2026)Li, Wu, Wang, Zhang, Zhu, Rossi, Morariu, and Kil]{hsl}
Zichao Li, Gang Wu, Zichao Wang, Ruiyi Zhang, Wanrong Zhu, Ryan~A. Rossi, Vlad~I Morariu, and Jihyung Kil.
\newblock Spinning straw into gold: Relabeling {LLM} agent trajectories in hindsight for successful demonstrations.
\newblock In \emph{The Fourteenth International Conference on Learning Representations}, 2026.

\bibitem[Liu et~al.(2025)Liu, Wang, Wu, Huang, Li, Zhang, and Jiao]{istar}
Xiaoqian Liu, Ke~Wang, Yuchuan Wu, Fei Huang, Yongbin Li, Junge Zhang, and Jianbin Jiao.
\newblock Agentic reinforcement learning with implicit step rewards.
\newblock \emph{arXiv preprint arXiv:2509.19199}, 2025.

\bibitem[Luo et~al.(2025)Luo, Shen, Yang, Zhao, Jwalapuram, Saha, Sahoo, Savarese, Xiong, and Li]{mcp-universe}
Ziyang Luo, Zhiqi Shen, Wenzhuo Yang, Zirui Zhao, Prathyusha Jwalapuram, Amrita Saha, Doyen Sahoo, Silvio Savarese, Caiming Xiong, and Junnan Li.
\newblock Mcp-universe: Benchmarking large language models with real-world model context protocol servers.
\newblock \emph{arXiv preprint arXiv:2508.14704}, 2025.

\bibitem[Lyu et~al.(2025)Lyu, Huang, Deng, Hoi, and An]{reloc}
Zhiyi Lyu, Jianguo Huang, Yanchen Deng, Steven Hoi, and Bo~An.
\newblock Let's revise step-by-step: A unified local search framework for code generation with llms.
\newblock \emph{arXiv preprint arXiv:2508.07434}, 2025.

\bibitem[{Notion}(2025)]{notion_ai}
{Notion}.
\newblock Notion: The ai workspace that works for you.
\newblock \url{https://www.notion.com/}, 2025.
\newblock Accessed: 2026-04-17.

\bibitem[Packer et~al.(2023)Packer, Fang, Patil, Lin, Wooders, and Gonzalez]{memGPT}
Charles Packer, Vivian Fang, Shishir\_G Patil, Kevin Lin, Sarah Wooders, and Joseph\_E Gonzalez.
\newblock Memgpt: towards llms as operating systems.
\newblock 2023.

\bibitem[Prabhakar et~al.(2025)Prabhakar, Liu, Zhu, Zhang, Awalgaonkar, Wang, Liu, Chen, Hoang, Niebles, et~al.]{apigen-mt}
Akshara Prabhakar, Zuxin Liu, Ming Zhu, Jianguo Zhang, Tulika Awalgaonkar, Shiyu Wang, Zhiwei Liu, Haolin Chen, Thai Hoang, Juan~Carlos Niebles, et~al.
\newblock Apigen-mt: Agentic pipeline for multi-turn data generation via simulated agent-human interplay.
\newblock \emph{arXiv preprint arXiv:2504.03601}, 2025.

\bibitem[Qi et~al.(2024)Qi, Liu, Iong, Lai, Sun, Zhao, Yang, Yang, Sun, Yao, et~al.]{webrl}
Zehan Qi, Xiao Liu, Iat~Long Iong, Hanyu Lai, Xueqiao Sun, Wenyi Zhao, Yu~Yang, Xinyue Yang, Jiadai Sun, Shuntian Yao, et~al.
\newblock Webrl: Training llm web agents via self-evolving online curriculum reinforcement learning.
\newblock \emph{arXiv preprint arXiv:2411.02337}, 2024.

\bibitem[Qin et~al.(2023)Qin, Liang, Ye, Zhu, Yan, Lu, Lin, Cong, Tang, Qian, et~al.]{toolllm}
Yujia Qin, Shihao Liang, Yining Ye, Kunlun Zhu, Lan Yan, Yaxi Lu, Yankai Lin, Xin Cong, Xiangru Tang, Bill Qian, et~al.
\newblock Toolllm: Facilitating large language models to master 16000+ real-world apis.
\newblock \emph{arXiv preprint arXiv:2307.16789}, 2023.

\bibitem[Rafailov et~al.(2023)Rafailov, Sharma, Mitchell, Manning, Ermon, and Finn]{dpo}
Rafael Rafailov, Archit Sharma, Eric Mitchell, Christopher~D Manning, Stefano Ermon, and Chelsea Finn.
\newblock Direct preference optimization: Your language model is secretly a reward model.
\newblock \emph{Advances in neural information processing systems}, 36:\penalty0 53728--53741, 2023.

\bibitem[Rawles et~al.(2024)Rawles, Clinckemaillie, Chang, Waltz, Lau, Fair, Li, Bishop, Li, Campbell-Ajala, et~al.]{androidworld}
Christopher Rawles, Sarah Clinckemaillie, Yifan Chang, Jonathan Waltz, Gabrielle Lau, Marybeth Fair, Alice Li, William Bishop, Wei Li, Folawiyo Campbell-Ajala, et~al.
\newblock Androidworld: A dynamic benchmarking environment for autonomous agents.
\newblock \emph{arXiv preprint arXiv:2405.14573}, 2024.

\bibitem[Schick et~al.(2023)Schick, Dwivedi-Yu, Dess{\`\i}, Raileanu, Lomeli, Hambro, Zettlemoyer, Cancedda, and Scialom]{toolformer}
Timo Schick, Jane Dwivedi-Yu, Roberto Dess{\`\i}, Roberta Raileanu, Maria Lomeli, Eric Hambro, Luke Zettlemoyer, Nicola Cancedda, and Thomas Scialom.
\newblock Toolformer: Language models can teach themselves to use tools.
\newblock \emph{Advances in neural information processing systems}, 36:\penalty0 68539--68551, 2023.

\bibitem[Shao et~al.(2024)Shao, Wang, Zhu, Xu, Song, Bi, Zhang, Zhang, Li, Wu, et~al.]{grpo}
Zhihong Shao, Peiyi Wang, Qihao Zhu, Runxin Xu, Junxiao Song, Xiao Bi, Haowei Zhang, Mingchuan Zhang, YK~Li, Yang Wu, et~al.
\newblock Deepseekmath: Pushing the limits of mathematical reasoning in open language models.
\newblock \emph{arXiv preprint arXiv:2402.03300}, 2024.

\bibitem[Shi et~al.(2024)Shi, Yuan, Wu, Wang, and Feng]{agentdpo}
Wentao Shi, Mengqi Yuan, Junkang Wu, Qifan Wang, and Fuli Feng.
\newblock Direct multi-turn preference optimization for language agents.
\newblock In \emph{Proceedings of the 2024 Conference on Empirical Methods in Natural Language Processing}, pages 2312--2324, 2024.

\bibitem[Su et~al.(2025)Su, Sun, Yoon, Yin, Yu, and Ar{\i}k]{learn_by_interact}
Hongjin Su, Ruoxi Sun, Jinsung Yoon, Pengcheng Yin, Tao Yu, and Sercan~{\"O} Ar{\i}k.
\newblock Learn-by-interact: A data-centric framework for self-adaptive agents in realistic environments.
\newblock \emph{arXiv preprint arXiv:2501.10893}, 2025.

\bibitem[Sun et~al.(2025)Sun, Cheng, Ding, Jin, Wang, Xu, Wu, Jia, Chen, Liu, et~al.]{os-genesis}
Qiushi Sun, Kanzhi Cheng, Zichen Ding, Chuanyang Jin, Yian Wang, Fangzhi Xu, Zhenyu Wu, Chengyou Jia, Liheng Chen, Zhoumianze Liu, et~al.
\newblock Os-genesis: Automating gui agent trajectory construction via reverse task synthesis.
\newblock In \emph{Proceedings of the 63rd Annual Meeting of the Association for Computational Linguistics (Volume 1: Long Papers)}, pages 5555--5579, 2025.

\bibitem[Tian et~al.(2026)Tian, Wang, Chen, Zhou, Yu, Zhang, Ouyang, Yin, Chen, Guo, et~al.]{astra}
Xiaoyu Tian, Haotian Wang, Shuaiting Chen, Hao Zhou, Kaichi Yu, Yudian Zhang, Jade Ouyang, Junxi Yin, Jiong Chen, Baoyan Guo, et~al.
\newblock Astra: Automated synthesis of agentic trajectories and reinforcement arenas.
\newblock \emph{arXiv preprint arXiv:2601.21558}, 2026.

\bibitem[Tu et~al.(2026)Tu, Hao, Yang, Chen, Zhang, Xia, Yang, Sun, Liu, Shen, et~al.]{scaleenv}
Dunwei Tu, Hongyan Hao, Hansi Yang, Yihao Chen, Yi-Kai Zhang, Zhikang Xia, Yu~Yang, Yueqing Sun, Xingchen Liu, Furao Shen, et~al.
\newblock Scaleenv: Scaling environment synthesis from scratch for generalist interactive tool-use agent training.
\newblock \emph{arXiv preprint arXiv:2602.06820}, 2026.

\bibitem[Wang et~al.(2025{\natexlab{a}})Wang, Dai, Ye, Gan, Yao, Deng, Wu, and Ying]{igpo}
Guoqing Wang, Sunhao Dai, Guangze Ye, Zeyu Gan, Wei Yao, Yong Deng, Xiaofeng Wu, and Zhenzhe Ying.
\newblock Information gain-based policy optimization: A simple and effective approach for multi-turn llm agents.
\newblock \emph{arXiv preprint arXiv:2510.14967}, 2025{\natexlab{a}}.

\bibitem[Wang et~al.(2025{\natexlab{b}})Wang, Hao, Dong, Zhang, Bao, Yang, and Wu]{ored}
Huaijie Wang, Shibo Hao, Hanze Dong, Shenao Zhang, Yilin Bao, Ziran Yang, and Yi~Wu.
\newblock Offline reinforcement learning for llm multi-step reasoning.
\newblock In \emph{Findings of the Association for Computational Linguistics: ACL 2025}, pages 8881--8893, 2025{\natexlab{b}}.

\bibitem[Wang et~al.(2026)Wang, Xu, Liu, Wang, Han, Yao, Yao, and He]{awm}
Zhaoyang Wang, Canwen Xu, Boyi Liu, Yite Wang, Siwei Han, Zhewei Yao, Huaxiu Yao, and Yuxiong He.
\newblock Agent world model: Infinity synthetic environments for agentic reinforcement learning.
\newblock \emph{arXiv preprint arXiv:2602.10090}, 2026.

\bibitem[Wu et~al.(2025)Wu, Liu, Zhang, Chen, Meng, Du, Zhao, Zhang, Ye, Wang, et~al.]{mcpmark}
Zijian Wu, Xiangyan Liu, Xinyuan Zhang, Lingjun Chen, Fanqing Meng, Lingxiao Du, Yiran Zhao, Fanshi Zhang, Yaoqi Ye, Jiawei Wang, et~al.
\newblock Mcpmark: A benchmark for stress-testing realistic and comprehensive mcp use.
\newblock \emph{arXiv preprint arXiv:2509.24002}, 2025.

\bibitem[Xi et~al.(2025)Xi, Liao, Li, Yang, Chen, Zhang, Wang, Jin, Zhou, Guan, et~al.]{agentprm}
Zhiheng Xi, Chenyang Liao, Guanyu Li, Yajie Yang, Wenxiang Chen, Zhihao Zhang, Binghai Wang, Senjie Jin, Yuhao Zhou, Jian Guan, et~al.
\newblock Agentprm: Process reward models for llm agents via step-wise promise and progress.
\newblock \emph{arXiv preprint arXiv:2511.08325}, 2025.

\bibitem[Xu et~al.(2024)Xu, Lu, Shen, Wang, Wang, Mao, Xiong, and Yu]{agenttrek}
Yiheng Xu, Dunjie Lu, Zhennan Shen, Junli Wang, Zekun Wang, Yuchen Mao, Caiming Xiong, and Tao Yu.
\newblock Agenttrek: Agent trajectory synthesis via guiding replay with web tutorials.
\newblock \emph{arXiv preprint arXiv:2412.09605}, 2024.

\bibitem[Xu et~al.(2025)Xu, Soria, Tan, Roy, Agrawal, Poovendran, and Panda]{toucan}
Zhangchen Xu, Adriana~Meza Soria, Shawn Tan, Anurag Roy, Ashish~Sunil Agrawal, Radha Poovendran, and Rameswar Panda.
\newblock Toucan: Synthesizing 1.5 m tool-agentic data from real-world mcp environments.
\newblock \emph{arXiv preprint arXiv:2510.01179}, 2025.

\bibitem[Yang et~al.(2025)Yang, Li, Yang, Zhang, Hui, Zheng, Yu, Gao, Huang, Lv, et~al.]{qwen3}
An~Yang, Anfeng Li, Baosong Yang, Beichen Zhang, Binyuan Hui, Bo~Zheng, Bowen Yu, Chang Gao, Chengen Huang, Chenxu Lv, et~al.
\newblock Qwen3 technical report.
\newblock \emph{arXiv preprint arXiv:2505.09388}, 2025.

\bibitem[Yao et~al.(2022)Yao, Zhao, Yu, Du, Shafran, Narasimhan, and Cao]{react}
Shunyu Yao, Jeffrey Zhao, Dian Yu, Nan Du, Izhak Shafran, Karthik~R Narasimhan, and Yuan Cao.
\newblock React: Synergizing reasoning and acting in language models.
\newblock In \emph{The eleventh international conference on learning representations}, 2022.

\bibitem[Yin et~al.(2024)Yin, Wang, Gu, Huang, Chen, and Zhou]{rpo}
Yueqin Yin, Zhendong Wang, Yi~Gu, Hai Huang, Weizhu Chen, and Mingyuan Zhou.
\newblock Relative preference optimization: Enhancing llm alignment through contrasting responses across identical and diverse prompts.
\newblock \emph{arXiv preprint arXiv:2402.10958}, 2024.

\bibitem[Yu et~al.(2025)Yu, Zhang, Zhu, Yuan, Zuo, Yue, Dai, Fan, Liu, Liu, et~al.]{dapo}
Qiying Yu, Zheng Zhang, Ruofei Zhu, Yufeng Yuan, Xiaochen Zuo, Yu~Yue, Weinan Dai, Tiantian Fan, Gaohong Liu, Lingjun Liu, et~al.
\newblock Dapo: An open-source llm reinforcement learning system at scale.
\newblock \emph{arXiv preprint arXiv:2503.14476}, 2025.

\bibitem[Yuan et~al.(2023)Yuan, Yuan, Li, Dong, Lu, Tan, Zhou, and Zhou]{rft}
Zheng Yuan, Hongyi Yuan, Chengpeng Li, Guanting Dong, Keming Lu, Chuanqi Tan, Chang Zhou, and Jingren Zhou.
\newblock Scaling relationship on learning mathematical reasoning with large language models.
\newblock \emph{arXiv preprint arXiv:2308.01825}, 2023.

\bibitem[Zhang et~al.(2025)Zhang, Lan, Zhu, Liu, Hoang, Kokane, Yao, Tan, Prabhakar, Chen, et~al.]{xlam}
Jianguo Zhang, Tian Lan, Ming Zhu, Zuxin Liu, Thai~Quoc Hoang, Shirley Kokane, Weiran Yao, Juntao Tan, Akshara Prabhakar, Haolin Chen, et~al.
\newblock xlam: A family of large action models to empower ai agent systems.
\newblock In \emph{Proceedings of the 2025 Conference of the Nations of the Americas Chapter of the Association for Computational Linguistics: Human Language Technologies (Volume 1: Long Papers)}, pages 11583--11597, 2025.

\bibitem[Zhou et~al.(2023)Zhou, Jiang, Cui, Wang, Xiao, Hou, Cotterell, and Sachan]{recurrentgpt}
Wangchunshu Zhou, Yuchen~Eleanor Jiang, Peng Cui, Tiannan Wang, Zhenxin Xiao, Yifan Hou, Ryan Cotterell, and Mrinmaya Sachan.
\newblock Recurrentgpt: Interactive generation of (arbitrarily) long text.
\newblock \emph{arXiv preprint arXiv:2305.13304}, 2023.

\end{thebibliography}

\appendix
\newpage

\section{Pseudo Code}
\label{app:algorithm}
We provide the complete AgentBrew pipeline in Algorithm~\ref{alg:agentbrew}. The procedure consists of three stages: raw experience collection (\S\ref{sec:experience collection}), experience distillation (\S\ref{sec:filter}), and PMI-weighted policy training (\S\ref{sec:training}). Stage 1 is the only phase that requires interaction with the live environment $\mathcal{E}$; Stages 2 and 3 proceed entirely offline using only the collected corpus and a reference language model $\pi_{\text{ref}}$.

\begin{algorithm}[H]
\caption{AgentBrew: Offline Tool-Use Agent Learning from Raw Trajectories}
\label{alg:agentbrew}
\begin{algorithmic}[1]
\REQUIRE Target environment $\mathcal{E}$ with tool set $\mathcal{F}$, base policy $\pi_\theta$, reference model $\pi_{\text{ref}}$, max turns $T_{\max}$
\ENSURE Trained policy $\pi_\theta$

\STATE \textbf{--- Stage 1: Raw Experience Collection ---}
\STATE Generate high-level task categories $\mathcal{G}$ from tool descriptions of $\mathcal{F}$
\FOR{each category $g_m \in \mathcal{G}$}
    \STATE Explore $\mathcal{E}$ via tool calls to retrieve live environmental information
    \STATE Propose grounded task instructions $\{I_n\}$ based on retrieved information
\ENDFOR
\FOR{each task $I_n \in \mathcal{I}$}
    \STATE Execute $I_n$ with $\pi_\theta$ in $\mathcal{E}$, producing trajectory $\tau_n = \{(s_t, a_t, o_t)\}_{t=0}^{T_n}$
    \IF{$T_n = T_{\max}$ (no natural termination)}
        \STATE Discard $\tau_n$
    \ENDIF
\ENDFOR
\STATE Collect raw corpus $\mathcal{D}_{\text{raw}} = \{(I_n, \tau_n)\}_{n=1}^{N}$ \hfill $\triangleright$ No quality filtering

\STATE
\STATE \textbf{--- Stage 2: Experience Distillation (Offline) ---}
\FOR{each $(I_n, \tau_n) \in \mathcal{D}_{\text{raw}}$}
    \STATE \textit{// Retrospective Task Inference}
    \STATE Remove chain-of-thought traces: $\bar{\tau}_n = \{(c_t, o_t)\}_{t=0}^{T_n}$
    \STATE Infer revised instruction: $\hat{I}_n \leftarrow \text{LLM}(\bar{\tau}_n)$ \hfill $\triangleright$ Goal-oriented, evidence-grounded
    \STATE
    \STATE \textit{// PMI-Based Credit Assignment}
    \STATE Compute unconditional NLL: $\mathcal{L}_{\emptyset} = -\log \pi_{\text{ref}}(\hat{I}_n)$
    \FOR{$t = 0, 1, \ldots, T_n$}
        \STATE Compute conditional NLL: $\mathcal{L}_t = -\log \pi_{\text{ref}}(\hat{I}_n \mid c_0, o_0, \ldots, c_t, o_t)$
    \ENDFOR
    \STATE $w_0 \leftarrow \mathcal{L}_{\emptyset} - \mathcal{L}_0$
    \FOR{$t = 1, \ldots, T_n$}
        \STATE $w_t \leftarrow \mathcal{L}_{t-1} - \mathcal{L}_t$ \hfill $\triangleright$ Per-action PMI credit
    \ENDFOR
    \STATE
    \STATE \textit{// Weight Normalization}
    \FOR{$t = 0, 1, \ldots, T_n$}
        \STATE $\tilde{w}_t \leftarrow 2 \cdot \max(w_t, 0) \;/\; \max_{0 \leq j \leq T_n} w_j$ \hfill $\triangleright$ Normalize to $[0, 2]$
    \ENDFOR
\ENDFOR

\STATE
\STATE \textbf{--- Stage 3: PMI-Weighted Policy Training (Offline) ---}
\STATE Optimize $\pi_\theta$ by minimizing:
\STATE \quad $\mathcal{J}(\theta) = -\displaystyle\sum_{n=1}^{N} \sum_{t=0}^{T_n} \tilde{w}_t^{(n)} \cdot \log \pi_\theta(a_t \mid \hat{I}_n, s_t)$

\end{algorithmic}
\end{algorithm}

\section{Environment Details}
\label{app:environments}

We evaluate AgentBrew on three real-world MCP applications that span distinct interaction paradigms: code collaboration (GitHub), knowledge management (Notion), and relational database administration (PostgreSQL). Table~\ref{tab:tool_sets} lists the complete tool set exposed by each environment.

\paragraph{GitHub.}
The GitHub environment exposes 41 tools covering the full spectrum of platform operations: repository management, file operations, issue tracking, pull request workflows, branch and commit inspection, CI/CD workflow management, code and user search, and security scanning (code scanning, Dependabot, secret scanning, and security advisories). Tasks in this environment require composing long action sequences across these tool categories. For example, forking a repository, creating a feature branch, pushing file changes, opening a pull request, and performing a code review, all within a single trajectory. The live state includes cross-referenced issues, pull requests, commits, and branches, making precise entity resolution critical.

\paragraph{Notion.}
The Notion environment provides 19 tools organized around four core object types: users, blocks, pages, and databases, along with search and comment capabilities. Despite the smaller tool count, Notion tasks are challenging because of deeply nested content structures: pages contain blocks that may themselves contain child blocks or inline database references, and databases define typed schemas with relations and rollups linking to other databases. Operations such as inserting a block at a precise position, restructuring a page layout, or synchronizing rows across related databases require the agent to maintain an accurate mental model of the workspace hierarchy throughout the interaction.

\paragraph{PostgreSQL.}
The PostgreSQL environment exposes 9 tools that combine standard SQL execution with database administration utilities. Beyond \texttt{execute\_sql} for arbitrary SQL statements, the environment provides schema inspection tools (\texttt{list\_schemas}, \texttt{list\_objects}, \texttt{get\_object\_details}), query analysis tools (\texttt{explain\_query}, \texttt{get\_top\_queries}), and health diagnostics (\texttt{analyze\_workload\_indexes}, \texttt{analyze\_query\_indexes}, \texttt{analyze\_db\_health}). Although the tool set is compact, the complexity arises from the SQL reasoning itself: tasks range from multi-table data migrations and schema design to query optimization and security policy implementation, requiring the agent to generate correct, multi-step SQL workflows grounded in the actual database schema and data.

\begin{table}[h]
\centering
\small
\caption{Tool sets exposed by each environment.}
\label{tab:tool_sets}
\begin{tabular}{@{}l c p{10.8cm}@{}}
\toprule
\textbf{Env.} & \textbf{\#Tools} & \textbf{Tools} \\
\midrule
GitHub & 41 & \scriptsize\texttt{create\_repository, get\_repository, list\_repositories, fork\_repository, get\_repository\_languages, get\_file\_contents, create\_or\_update\_file, delete\_file, push\_files, create\_issue, get\_issue, list\_issues, update\_issue, add\_issue\_comment, create\_pull\_request, get\_pull\_request, list\_pull\_requests, get\_pull\_request\_files, create\_pull\_request\_review, merge\_pull\_request, create\_branch, list\_branches, list\_commits, list\_workflow\_runs, get\_workflow\_run, list\_workflow\_jobs, cancel\_workflow\_run, search\_repositories, search\_code, search\_issues, search\_users, list\_code\_scanning\_alerts, get\_code\_scanning\_alert, list\_dependabot\_alerts, get\_dependabot\_alert, list\_secret\_scanning\_alerts, get\_secret\_scanning\_alert, list\_global\_security\_advisories, list\_org\_repository\_security\_advisories, list\_repository\_security\_advisories, create\_pull\_request\_with\_copilot} \\
\midrule
Notion & 19 & \scriptsize\texttt{get-user, get-users, get-self, post-database-query, post-search, get-block-children, patch-block-children, retrieve-a-block, update-a-block, delete-a-block, retrieve-a-page, patch-page, post-page, create-a-database, update-a-database, retrieve-a-database, retrieve-a-page-property, retrieve-a-comment, create-a-comment} \\
\midrule
Postgres & 9 & \scriptsize\texttt{list\_schemas, list\_objects, get\_object\_details, execute\_sql, explain\_query, get\_top\_queries, analyze\_workload\_indexes, analyze\_query\_indexes, analyze\_db\_health} \\
\bottomrule
\end{tabular}
\end{table}

\section{Task Proposal}
\label{app:task_proposal}

\paragraph{Environment preparation.}
Each target environment $\mathcal{E}$ requires a diverse substrate of real-world content to ensure that proposed tasks are grounded in meaningful state. For \textbf{GitHub}, we curate 200 public repositories spanning a variety of programming languages (Python, JavaScript, Go, Rust, etc.) and project types (web applications, CLI tools, data pipelines, machine learning libraries), providing a rich landscape of issues, pull requests, branches, and commit histories for the agent to interact with. For \textbf{Notion}, we collect 180 free templates from the official Notion template gallery,\footnote{\url{https://www.notion.com/templates}} covering domains such as project management, personal planning, knowledge bases, and team collaboration. These templates provide pre-populated pages, databases, and relational structures that mirror realistic workspace configurations. For \textbf{PostgreSQL}, we assemble 145 relational databases sourced from established open-source benchmarks including BIRD \citep{bird} and Spider \citep{spider}, which collectively span domains such as retail, healthcare, education, sports, and enterprise operations. This diversity ensures that the subsequent task proposal stage can generate instructions grounded in entities and configurations that actually exist within each environment.

\paragraph{High-level task category generation.}
Given the tool set $\mathcal{F}$ of the target environment $\mathcal{E}$, we prompt the agent to generate a set of high-level task categories $\mathcal{G} = \{g_1, g_2, \ldots, g_M\}$ that cover representative human intents supported by $\mathcal{E}$. The agent receives the names and descriptions of all available tools as input and produces categories that reflect common multi-step usage patterns, each requiring the composition of multiple tools. Table~\ref{tab:task_categories} lists the generated categories for each environment. GitHub yields 14 categories spanning repository lifecycle management, code collaboration, and project planning; Notion produces 12 categories covering database operations, page editing, and content synthesis; PostgreSQL results in 15 categories ranging from data migration and analytics to security policy design. The prompt template used for category generation is shown below.
\begin{tcolorbox}[
  colback=gray!5,
  colframe=sapphireblue,
  boxrule=0.5pt,
  arc=2pt,
  left=6pt, right=6pt, top=6pt, bottom=6pt,
  title={\small\textbf{Prompt: Task Category Generation}},
  fonttitle=\small
]
\small
You are an expert task designer for evaluating AI agents that interact with real-world applications via tool-use APIs.

You are given the complete tool set of the \texttt{\{environment\_name\}} environment. Each tool is described by its name, purpose, and parameter specification:

\texttt{\{tool\_descriptions\}}

Your goal is to generate a diverse set of high-level task categories that represent realistic human intents on this platform. Each category should:

(1) Require \textbf{composing multiple tools} in a meaningful sequence (not single-tool invocations).

(2) Reflect a \textbf{concrete, goal-oriented workflow} that a real user would perform (e.g., ``migrate data between tables with validation'', not ``use the search API'').

(3) Be \textbf{distinct} from other categories in terms of the tool combinations and reasoning patterns involved.

(4) Cover \textbf{diverse difficulty levels}, from straightforward CRUD operations to complex multi-step workflows involving conditional logic, cross-entity coordination, or synthesis.

\medskip
For each category, provide:

-- \textbf{Title}: A concise name for the category.

-- \textbf{Goal}: A one-sentence description of what a user aims to accomplish.

\medskip
Generate \texttt{\{num\_categories\}} categories that collectively maximize the coverage of the tool set.
\end{tcolorbox}

\begin{table}[h]
\centering
\small
\caption{Task categories generated for each environment.}
\label{tab:task_categories}
\resizebox{\textwidth}{!}{%
\setlength{\tabcolsep}{4pt}
\begin{tabular}{@{}rl@{\hskip 10pt}rl@{\hskip 10pt}rl@{}}
\toprule
\multicolumn{2}{c}{\textbf{GitHub (14)}} & \multicolumn{2}{c}{\textbf{Notion (12)}} & \multicolumn{2}{c}{\textbf{PostgreSQL (15)}} \\
\midrule
1. & Project bootstrap \& self-review        & 1. & Database creation \& schema design        & 1. & Bulk migration \& transformation \\
2. & Multi-source code integration            & 2. & Cross-database relation \& sync           & 2. & Hierarchy-aware CRUD \& integrity \\
3. & Ecosystem metrics \& reporting           & 3. & Batch query, filter \& migration          & 3. & Dashboard reporting \& aggregates \\
4. & Decision-based comparative forking       & 4. & Batch updates \& coordinated edits        & 4. & SQL query debugging \& fixing \\
5. & Simulated issue automation               & 5. & Aggregate computation \& dashboards       & 5. & Query optimization \& indexing \\
6. & Fork-and-fix contribution                & 6. & Multi-source query \& synthesis           & 6. & Operational workflow management \\
7. & Cross-repo dependency injection          & 7. & Page layout \& column restructuring       & 7. & Descriptive reporting \& statistics \\
8. & Epic \& sub-issue planning               & 8. & Block insertion \& format matching         & 8. & Multi-factor performance analytics \\
9. & Release audit \& upgrade docs            & 9. & Template filling \& nested editing        & 9. & Schema buildout \& seed data \\
10. & Repo maintenance \& cleanup             & 10. & Content deletion \& archival             & 10. & Retention \& churn analysis \\
11. & Code archaeology \& arch.\ docs         & 11. & Style \& color batch updates             & 11. & Executive dashboard automation \\
12. & Legacy branch backporting               & 12. & Itinerary \& planner composition         & 12. & Structural \& bottleneck analysis \\
13. & Research framework scaffolding          &     &                                           & 13. & Consistency enforcement \\
14. & PR queue management \& hygiene          &     &                                           & 14. & Security policy \& row-level access \\
    &                                          &     &                                           & 15. & Transactional procedures \& audit \\
\bottomrule
\end{tabular}%
}
\end{table}

\paragraph{Grounded task instruction generation.}
For each category $g_m \in \mathcal{G}$, a task generation agent conducts a lightweight exploratory interaction with $\mathcal{E}$ to ground the abstract category in concrete environmental state. The agent follows a ReAct-style loop: it first invokes a single read-only tool call (e.g., listing repository contents, querying database schemas, or browsing page structures) to discover real entities such as repository names, file paths, issue numbers, page IDs, or table columns. Based on the retrieved information, the agent then formulates a specific task instruction $I_n$ written as a natural user request that references only verified identifiers. This grounding step is critical because tasks that reference nonexistent resources (e.g., a deleted branch or a missing database entry) would produce unexecutable trajectories during the subsequent data collection phase. To maintain diversity, the agent also receives an example task from a different category as a style reference, ensuring consistent tone and specificity across the generated task set. The prompt template is shown below.

\begin{tcolorbox}[
  colback=gray!5,
  colframe=sapphireblue,
  boxrule=0.5pt,
  arc=2pt,
  left=6pt, right=6pt, top=6pt, bottom=6pt,
  title={\small\textbf{Prompt: Grounded Task Instruction Generation}},
  fonttitle=\small
]
\small
You are a ReAct (Reasoning and Acting) agent specializing in Task Engineering. You are designing a single, high-quality task for LLMs that interact with real-world applications via tool-use.

\medskip
\textbf{Tool Restrictions:} You may ONLY call read/search tools to inspect the target environment --- do NOT create, modify, or delete anything. Call exactly 1 tool to ground yourself, then immediately output the final task.

\medskip
\textbf{What the generated task CAN include:} The task may involve any operation supported by the environment (reading, creating, updating, deleting resources). Write operations must target user-owned resources (newly created or forked).

\medskip
\textbf{Your Job:}

(1) Read the \textbf{Workflow Pattern} (\texttt{\{task\_category\}}) to understand the type of task to generate.

(2) Read the \textbf{Example Task} (\texttt{\{example\_task\}}) to learn the desired tone, style, and level of detail.

(3) Call 1 read-only tool to inspect the target environment and discover concrete entities.

(4) Write a NEW task that follows the same workflow pattern, mimics the example's tone, and is grounded in the environment's real state.

\medskip
\textbf{Available tools:} \texttt{\{tool\_descriptions\}}

\medskip
\textbf{Task Output Constraints:}

(1) Output exactly one task. Do not generate multiple tasks or sub-task lists.

(2) Write it as a natural user request --- same tone and style as the example. No formal specs, no numbered checklists, no headings.

(3) Do not mention ``tools'' or ``APIs'' in the final task text.

(4) Include concrete identifiers confirmed via the tool call.

\medskip
\textbf{Output Format:} A JSON object with a \texttt{thought} field (verification reasoning) and either an \texttt{action} field (to call a read-only tool) or an \texttt{answer} field (the final natural-language task instruction).
\end{tcolorbox}

\section{PMI-Based Credit Assignment}
\label{app:pmi-credit-details}
For each trajectory, we first infer a hindsight instruction $\hat{I}$ that describes the outcome actually supported by the observed tool interactions. We then compute per-action credit by measuring how much each trajectory prefix reduces the negative log-likelihood of $\hat{I}$ under a reference model $\pi_{\mathrm{ref}}$.
Let $\bar{\tau}_{\le t}$ denote the serialized tool-interaction prefix up to step $t$, and let $\hat{I} = (\hat{i}_1,\ldots,\hat{i}_m)$ be the tokenized hindsight instruction. In implementation, we compute the average token negative log-likelihood:
\[
L_t
=
-\frac{1}{m}
\sum_{k=1}^{m}
\log
\pi_{\mathrm{ref}}
\left(
\hat{i}_k
\mid
\hat{i}_{<k}, \bar{\tau}_{\le t}
\right).
\]
The empty-prefix loss is denoted by $L_{-1}$ and is computed in the same way without conditioning on any observed tool interactions:
\[
L_{-1}
=
-\frac{1}{m}
\sum_{k=1}^{m}
\log
\pi_{\mathrm{ref}}
\left(
\hat{i}_k
\mid
\hat{i}_{<k}
\right).
\]
The raw PMI credit for action $t$ is the reduction in hindsight-instruction NLL after revealing that action:
\[
w_t = L_{t-1} - L_t .
\]
Thus, positive credit indicates that the action makes the inferred instruction easier to predict, while zero or negative credit indicates that the action is redundant or uninformative with respect to the inferred instruction.
Before training, we clamp negative credits and normalize weights within each trajectory:
\[
\tilde{w}_t
=
2 \cdot
\frac{\max(w_t,0)}
{\max_j w_j}.
\]
If all raw credits in a trajectory are non-positive, we set all normalized weights to zero.

\paragraph{Empirical statistics of normalized weights.}
Across the full training corpus, the mean normalized weight $\tilde{w}_t$ is 0.75, with per-environment values of 0.81 (GitHub), 0.71 (Notion), and 0.77 (PostgreSQL).
This value is below the uniform baseline of 1.0 because negative credits are clamped to zero and many actions within each trajectory receive near-zero credit, reflecting the intended selective effect of PMI weighting.
The relative ranking of per-action credits, rather than the absolute scale, is the primary mechanism through which PMI weighting steers the policy toward informative actions.

\section{Retrospective Task Inference}
\label{app:prompt_inference}

As described in \S\ref{sec:realignment}, we infer a revised instruction $\hat{I}_n$ from each trajectory's reduced form $\bar{\tau}_n = \{(c_t, o_t)\}_{t=0}^{T_n}$, which retains only tool calls and environment responses. We use Qwen3-32B for this inference, consistent with the rest of the AgentBrew pipeline. Before prompting, we extract a structured change summary from $\bar{\tau}_n$ by identifying successful state-changing operations (e.g., pages created, rows inserted, SQL objects defined) and separating them from read-only or failed steps. Both the change summary and the full reduced trajectory are provided as input. 

The prompt follows a consistent template across all three environments, instantiated with environment-specific terminology and examples. The general template is shown below.

\begin{tcolorbox}[
  colback=gray!5,
  colframe=sapphireblue,
  boxrule=0.5pt,
  arc=2pt,
  left=6pt, right=6pt, top=6pt, bottom=6pt,
  title={\small\textbf{Prompt: Retrospective Task Inference}},
  fonttitle=\small
]
\small
You revise benchmark-style \texttt{\{environment\_name\}} tasks from agent tool trajectories.

Your job is to minimally edit the ORIGINAL TASK so it matches what the agent actually completed.

\textbf{Main goal:}
\begin{itemize}
\item Remove or weaken unsupported subtasks.
\item The revised task must be highly consistent with the trajectory.
\end{itemize}

Use the trajectory as evidence, with priority on: (1) successful state-changing operations; (2) query/retrieve steps only when they help identify objects involved in successful writes.

\textbf{Rules:}
\begin{itemize}
\item Output only the revised task text.
\item Preserve the original high-level task type (e.g., migration stays migration, analysis stays analysis).
\item If the agent only completed part of a multi-step task, keep only the completed subset.
\item Do not claim outcomes (e.g., performance improvements, verification success) unless directly supported.
\item Do not invent entities, artifacts, or results not evidenced by the trajectory.
\item Prefer deleting unsupported details over replacing them with new speculative ones.
\item Do not preserve strong claims (\texttt{all}, \texttt{each}, \texttt{exact count}, \texttt{verbatim}) unless directly supported by successful operations.
\end{itemize}

\textbf{ORIGINAL TASK:}\\
\texttt{\{original\_task\}}

\textbf{OBSERVED CHANGES:}\\
\texttt{\{change\_summary\}}

\textbf{TRAJECTORY:}\\
\texttt{\{trajectory\_str\}}

\end{tcolorbox}

\section{Experiment Details}
\label{app:implementation}

\subsection{Context Management}
\label{app:context}

Real-world tool-use interactions frequently produce trajectories exceeding 100K tokens, as each tool call may return verbose API responses (e.g., full page content in Notion or lengthy query results in PostgreSQL). To handle this within finite context windows, we adopt a sliding-window memory mechanism inspired by prior work on memory-augmented agents \citep{memGPT,recurrentgpt}.

At each step $t$, the agent's context is composed of two parts: (1)~a \emph{long-term memory} consisting of compressed summaries of all steps prior to the recent window, and (2)~a \emph{short-term context} containing the full, uncompressed content of the three most recent steps $(s_{t-2}, a_{t-2}, o_{t-2}), \ldots, (s_t, a_t, o_t)$. This design preserves fine-grained details for immediate decision-making while retaining high-level progress information from earlier steps.

Concretely, after each step $t$, we prompt the LLM to generate a concise summary of the current step conditioned on the task instruction, all previous step summaries, and the full content of step $t$. The summary captures what was attempted, key results or facts discovered, important constraints or decisions, and concrete identifiers (paths, URLs, IDs). Each summary is kept under 500 words and is appended to the long-term memory buffer. When constructing the context for step $t{+}1$, all summaries for steps $0$ to $t$ are concatenated as the long-term memory prefix, followed by the full content of steps $t{-}2$, $t{-}1$, and $t$.

This mechanism is applied uniformly across all methods and models in our experiments, ensuring that performance differences reflect the training approach rather than context management advantages. The prompt template used for step summarization is provided below.

\begin{tcolorbox}[
  colback=gray!5,
  colframe=sapphireblue,
  boxrule=0.5pt,
  arc=2pt,
  left=6pt, right=6pt, top=6pt, bottom=6pt,
  title={\small\textbf{Prompt: Step-Level Summarization}},
  fonttitle=\small
]
\small
You are a step-level summarizer for a multi-step ReAct-style agent.

\textbf{Your task:} Given the overall question, all past step summaries, and the full content of the current step, extract only the information from this step that will be useful for understanding what has already been done and making better decisions in future steps.

\medskip
\textbf{Overall Question:} \texttt{\{question\}}

\textbf{Past Step Summaries:} \texttt{\{past\_summaries\}}

\textbf{Current Step Index:} \texttt{\{step\_index\}}

\textbf{Full Content of the Current Step:} \texttt{\{step\_full\_content\}}

\medskip
Produce a concise summary of this current step only, following these rules:
\begin{itemize}
\item Focus on: what was attempted or executed; key results or facts discovered; important constraints, assumptions, or decisions; identifiers, paths, or URLs needed later.
\item Connect to the question and past progress when relevant, but do not re-summarize previous steps.
\item Omit generic reasoning boilerplate, tool framework noise, and large raw outputs (summarize their essential points instead).
\item Use fewer than 500 words.
\end{itemize}
\end{tcolorbox}

\subsection{Training Hyperparameters}
\label{app:training}

We train the policy $\pi_\theta$ using the PMI-weighted SFT objective (Eq.~\ref{eq:objective}) implemented as a custom weighted cross-entropy loss. For each training sample, the per-action credit $\tilde{w}_t^{(n)}$ is attached as a sample-level weight; during forward passes, the trainer computes per-sample token-averaged cross-entropy and scales it by the corresponding weight before aggregating across the batch. During evaluation, weights are disabled to ensure comparable eval loss across runs.

We fine-tune Qwen3-32B using DeepSpeed ZeRO-Stage 3 on 4$\times$H100 80GB GPUs with the following configuration: learning rate $5 \times 10^{-6}$ with cosine scheduling and 10\% warmup, per-device batch size 2 with gradient accumulation over 4 steps (effective batch size 64), weight decay 0.1, maximum sequence length 20{,}000 tokens, BF16 mixed precision, and gradient checkpointing enabled. Training runs for 1 epoch over the combined corpus of all three environments. Samples with zero normalized credit ($\tilde{w}_t^{(n)} = 0$) are filtered out before training, as they contribute no gradient signal. The 14B transfer experiments (\S\ref{sec:analysis_transfer}) use identical hyperparameters and training data.

\subsection{Evaluation Protocol}
\label{app:evaluation}

We adopt the evaluation framework from MCP-Universe \citep{mcp-universe}. Each test task is paired with a programmatic verifier that inspects the post-execution environment state (e.g., created resources, modified database rows, updated page content) against a ground-truth specification. The verifier produces a per-task score in $[0, 1]$ reflecting partial completion, and \textbf{Acc} is the fraction of tasks achieving a score of $1.0$ (full completion). \textbf{Score} is the average per-task score across all test tasks.


\section{Case Studies}
\label{app:case_studies}

We provide qualitative case studies across three real-world tool-use environments:
GitHub, Notion, and PostgreSQL. These examples illustrate how AgentBrew distills
useful supervision from imperfect raw trajectories. In each case, retrospective
task inference rewrites the original instruction into an aligned task that better
matches the trajectory's observable outcome, while PMI credit
assignment identifies which actions are useful for policy learning.

\begin{figure}[p]
    \centering
    \includegraphics[
        width=\linewidth,
        height=0.88\textheight,
        keepaspectratio
    ]{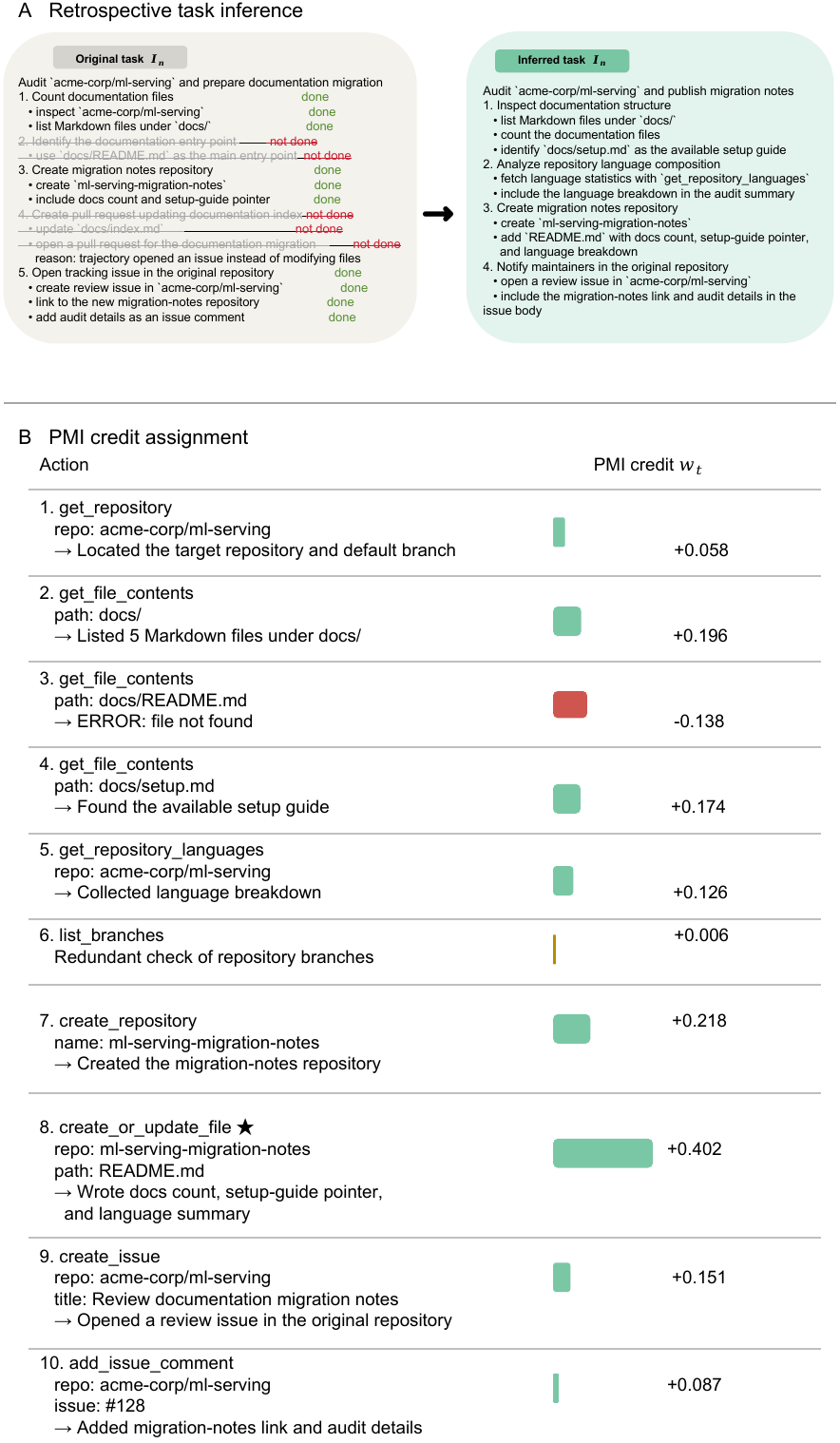}
    \caption{Case study on the GitHub environment.}
    \label{fig:case_github}
\end{figure}

\begin{figure}[p]
    \centering
    \includegraphics[
        width=\linewidth,
        height=0.88\textheight,
        keepaspectratio
    ]{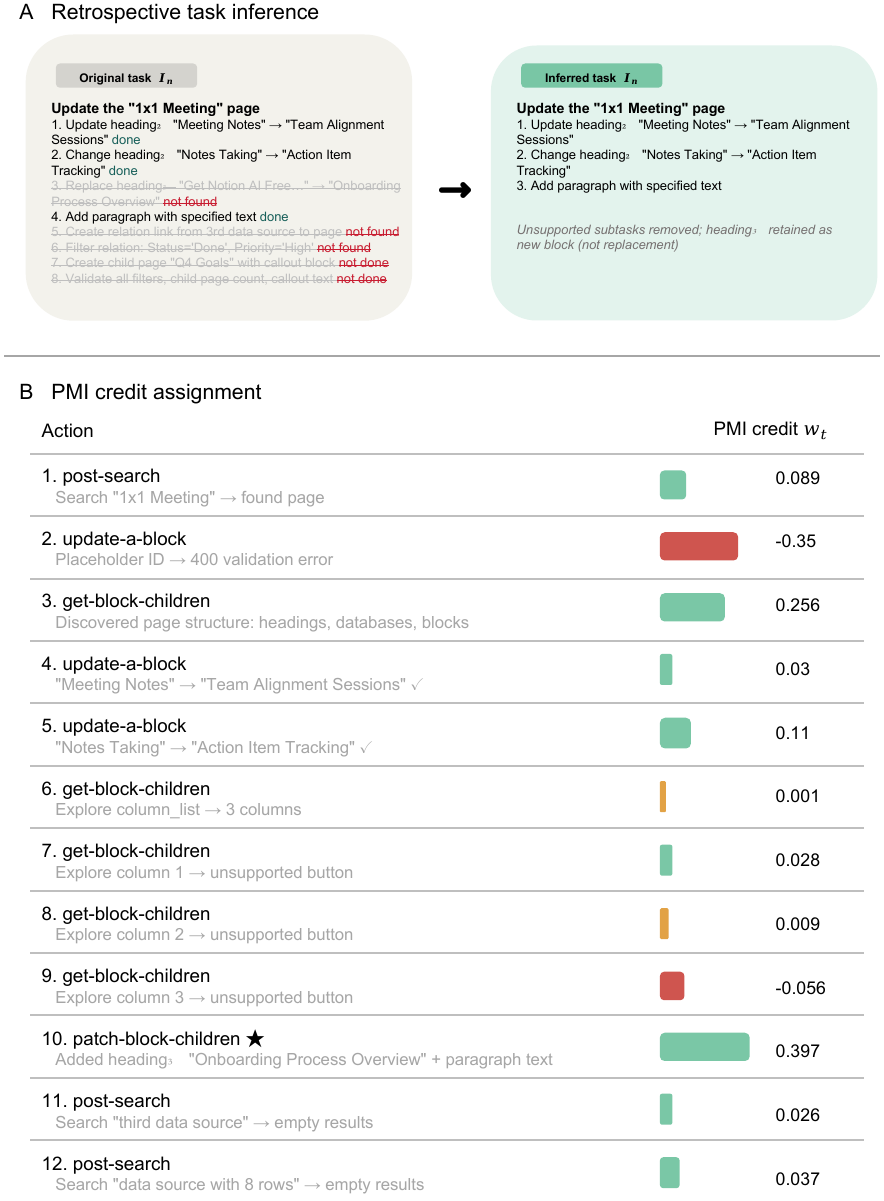}
    \caption{Case study on the Notion environment.}
    \label{fig:case_notion}
\end{figure}

\begin{figure}[p]
    \centering
    \includegraphics[
        width=\linewidth,
        height=0.88\textheight,
        keepaspectratio
    ]{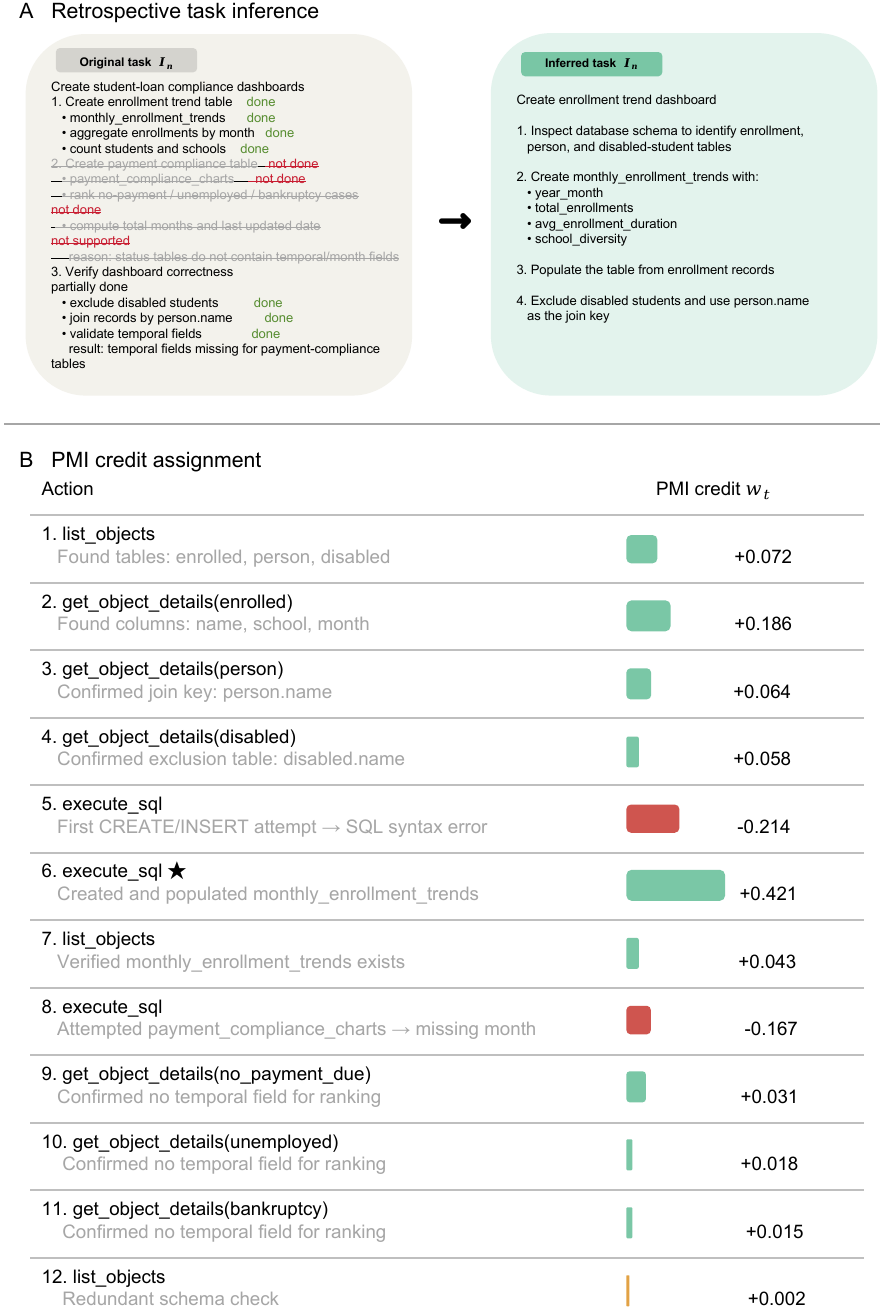}
    \caption{Case study on the PostgreSQL environment.}
    \label{fig:case_postgresql}
\end{figure}


\end{document}